\documentclass[10pt,twocolumn,letterpaper]{article}
\usepackage[margin=0.72in,columnsep=0.25in]{geometry}
\usepackage[T1]{fontenc}
\usepackage{times}
\usepackage{amsmath,amssymb}
\usepackage{graphicx}
\usepackage{booktabs,tabularx,multirow,makecell,array}
\usepackage[table]{xcolor}
\usepackage{tikz}
\usetikzlibrary{positioning,arrows.meta,fit,backgrounds,calc,shapes.misc}
\usepackage[numbers,sort&compress]{natbib}
\usepackage{enumitem}
\usepackage{caption}
\usepackage{placeins}
\usepackage[hidelinks]{hyperref}
\usepackage{xurl}
\usepackage[capitalise,noabbrev]{cleveref}

\setlist{nosep,leftmargin=*}
\newcommand{\tabsize}{\scriptsize}
\newcolumntype{L}[1]{>{\raggedright\arraybackslash}p{#1}}
\newcolumntype{Y}{>{\raggedright\arraybackslash}X}
\definecolor{boxblue}{HTML}{2a78d6}
\definecolor{boxorange}{HTML}{eb6834}
\definecolor{boxgreen}{HTML}{1baf7a}
\definecolor{boxviolet}{HTML}{6250d6}
\definecolor{boxgray}{HTML}{52514e}
\newcommand{\takeaway}[1]{\par\smallskip\noindent\fcolorbox{boxgray!40}{boxgray!6}{\parbox{\dimexpr\columnwidth-2\fboxsep-2\fboxrule}{\small\textbf{Takeaway.} #1}}\par\smallskip}

\title{\Large\bfseries Dexterous Robotic Hands in the Foundation-Model Era:\\ A Full-Stack Survey of Hardware, Data, Cross-Embodiment Transfer, Learning, and Evaluation}
\author{Ji~Pei\thanks{Corresponding author. Email: \texttt{ji.pei@stu.xjtu.edu.cn}} \\[2pt]
\normalsize Xi'an Jiaotong University, Xi'an, China}
\date{}

\begin{document}
\raggedbottom
\maketitle

\begin{abstract}
\noindent
Dexterous robotic hands are increasingly coupled to large-scale human data and foundation models, yet progress in hardware, representation learning and evaluation is often assessed separately. This survey reviews work from January 2024 to early November 2026, with earlier studies included to establish technical lineages. We organize the literature into five interdependent layers: hardware; data collection and curation; cross-embodiment transfer; learning and foundation models; and evaluation and applications. Human hand--object understanding and tactile sensing provide two cross-cutting perspectives. A system-level architecture, comparative tables and five case analyses connect design choices across these layers. We distinguish within-study experimental comparisons from heterogeneous corpus statistics, vendor specifications and company-reported results. Three conclusions emerge. First, the value of human data depends on kinematic, visual and contact alignment, not recording volume alone. Second, shared action representations and foundation policies do not remove embodiment-specific constraints on force, latency and contact. Third, task completion is insufficient to establish physical correctness or deployment readiness. Evidence remains limited for zero-shot transfer of contact-rich skills between morphologically different hands, long-duration reliability, and cross-sensor tactile generalization. We identify eleven open problems and propose reporting interfaces and controlled comparisons that can make full-stack progress more measurable. The review is a structured narrative synthesis rather than a pooled performance ranking or an exhaustive systematic review.
\end{abstract}


\section{Introduction}
\label{sec:intro}

The human hand is the interface through which most of the built world is operated: tools, containers, fasteners, cables, keyboards and laboratory glassware are all designed around five fingers, opposable thumbs and a sense of touch. Building robot hands that can use this world without redesigning it has been a goal of robotics for decades~\cite{okamura,piazza,controzzi,billard}. For most of that time, progress was limited on three fronts at once. Anthropomorphic hands were expensive, fragile and hard to maintain (the Shadow Dexterous Hand~\cite{shadowhand} costs more than US\$100k); data for multi-fingered manipulation were scarce because teleoperating 20 joints is slow and lacks force feedback; and learning methods for high-DoF hands were largely confined to single tasks trained with reinforcement learning (RL) in simulation~\cite{dactyl,rubik,dextreme,visualdexterity}.

Between 2024 and 2026 all three constraints loosened at the same time. On the \emph{hardware} side, a 22-DoF hand became available for about US\$6.5k~\cite{unitreedex5s}, a six-DoF linkage hand for under US\$1k~\cite{icra26hands}, and open-source hands for a few hundred dollars~\cite{aerohand,crafthand}; Chinese market trackers report shipments growing from about 19k units in 2025 to a forecast of 70k in 2026~\cite{ggii}. On the \emph{data} side, collection moved away from teleoperating a robot towards capturing humans directly: wearable exoskeletons and gloves that are kinematically matched to the robot hand~\cite{dexumi,dexop,dexwild,realdexumi}, and egocentric video with 3D hand tracking at the scale of $10^3$--$10^5$ hours~\cite{egodex,egoscale,beingh05,dreamdojo}. Industrial players report even larger corpora, with press reports of 270k to more than 500k hours of wearable data~\cite{gen15,generalistprofile} and about ten million glove episodes collected in homes~\cite{sunday36kr}. On the \emph{learning} side, generalist VLA models~\cite{pi0,pi05,pi07,gr00tn1,geminirobotics} and world-action models~\cite{dreamzero,beingh07} began to target high-DoF hands, with hierarchical systems that add a dedicated high-frequency contact layer below the policy~\cite{figurehelix02,craftnet}.

These developments are tightly coupled. A hand whose kinematics and camera placement mirror a human's can reuse human data with little retargeting error; data collected with a hand-specific exoskeleton usually require additional retargeting for reuse with a different hand; whether a unified action space is needed depends on how many embodiments a model must serve; and benchmark results only transfer if hardware, sensing and evaluation protocols are comparable. Recent evidence makes this coupling explicit. EgoScale reports a log-linear relation between human-pretraining hours and validation loss, alongside improved downstream completion~\cite{egoscale}; Liber-0 reports that, under identical post-training, 10k hours of glove data outperform 2M hours of web video, and that a more human-like robot benefits more from the same human data~\cite{liber0}; DexWild finds that too much human data relative to robot data (1:5) hurts performance compared with 1:2~\cite{dexwild}. None of these findings can be understood from the data layer alone.

\Cref{fig:timeline} groups selected milestones by year and layer.

\Cref{fig:fullstack} gives the complementary systems view: the survey follows the forward path from human or robot data to deployment, while treating failures, contact signals and benchmark results as feedback into the next design cycle.

\paragraph{Scope and positioning.} Classical surveys cover the mechanics of robot hands~\cite{piazza,controzzi}, dexterous manipulation in general~\cite{okamura,billard}, or a single learning paradigm such as RL~\cite{dexrlreview}, in-hand manipulation~\cite{weinberg} or imitation learning~\cite{dexilsurvey}. A recent survey organizes embodied data into a ``pyramid''~\cite{datapyramid} without focusing on hands. In contrast, we treat the dexterous hand as a \emph{full stack} and survey how its layers co-evolve. We focus on work released between January 2024 and early November 2026, and include earlier work where it defines a lineage (e.g., Dactyl~\cite{dactyl} for sim-to-real in-hand reorientation, DexPilot~\cite{dexpilot} for vision-based retargeting).

\paragraph{Contributions.}
\begin{itemize}
  \item A five-layer taxonomy (hardware, data, cross-embodiment transfer, learning, evaluation) that makes the dependencies between layers explicit (\cref{sec:taxonomy}).
  \item A structured review of each layer with comparison tables covering more than 40 hands, more than 50 data-collection systems and datasets, human hand--object datasets, more than 30 cross-embodiment methods, tactile models, grasp-synthesis methods, about 20 industrial foundation-model programs and more than 20 benchmarks (\cref{sec:hardware,sec:data,sec:human,sec:crossemb,sec:touch,sec:learning,sec:eval}).
  \item Two cross-cutting analyses: human hands as a data source (\cref{sec:human}) and touch across the stack (\cref{sec:touch}).
  \item A quantitative synthesis of reported evidence, including source-separated data scales, within-study collection ratios, price-versus-DoF comparisons and human-data ablations (\cref{fig:datascale,fig:throughput,fig:pricedof}).
  \item Five full-stack case studies, including a deployment design for a commercial kitchen, that trace how decisions propagate across layers (\cref{sec:cases}).
  \item Eleven open problems and concrete recommendations for researchers and practitioners (\cref{sec:challenges}).
\end{itemize}

\paragraph{Review design and coverage.} This is a structured narrative review of sources released between January 2024 and 4 November 2026, with earlier work included for foundational methods. The source collection covers arXiv, major robotics, learning and vision venues, project pages, company technical blogs and trade press. Core studies concern multi-fingered hands; adjacent gripper and whole-body studies are included when they motivate a transferable component and are identified as such. Although reference-checking notes accompany the source collection, a complete search log, screening counts and an independent dual-review record are not available. We therefore do not claim exhaustive coverage or systematic-review completeness. Publication, language, company-disclosure and positive-result biases may affect the synthesis.

\paragraph{Evidence and comparability.} We separate the source of a claim (paper/preprint, company technical report, or secondary press) from the setting in which it was measured (simulation, real robot, or deployment). These are distinct axes: a large real-world result can still be company-reported, and peer review does not make two benchmarks comparable. Reported percentages retain their original success or progress semantics; missing trial counts, error bars and uncertainty cannot be reconstructed from headline scores. Within-study ablations support local comparisons, whereas cross-paper plots summarize reported scale or cost and are not leaderboards. Recommended architectures, maturity assessments and the kitchen design are explicitly our synthesis, not independently validated deployments.

\section{A Full-Stack Taxonomy}
\label{sec:taxonomy}

\begin{figure*}[t]
\centering
\includegraphics[width=\textwidth]{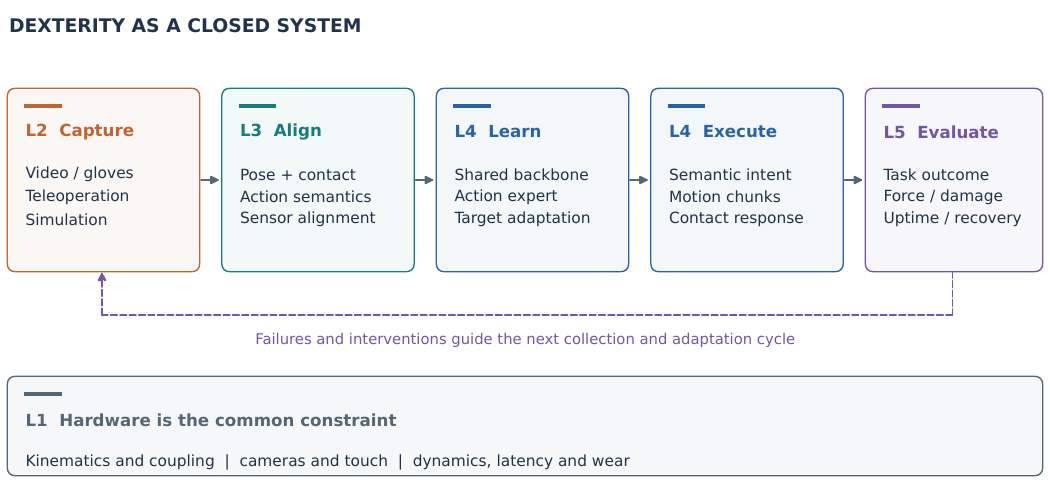}
\caption{The five-layer stack as a closed system. Hardware constrains capture, alignment and execution; evaluation informs the next data and adaptation cycle. This is a conceptual synthesis of the reviewed literature, not the implementation of a particular product. The arrows indicate dependencies rather than guaranteed causal improvements.}
\label{fig:fullstack}\label{fig:taxonomy}
\end{figure*}

\Cref{fig:fullstack} organizes the survey around three questions: what information survives the mapping from a human to a robot; which decisions can be shared across hands; and which failures require feedback close to the hardware? The five layers answer these questions jointly. Human data are useful only after their observations and action semantics are aligned; a shared policy still needs feasible target-hand commands; and a high task score does not establish reliable contact or deployment readiness.

\paragraph{Interfaces rather than isolated components.} The minimum information at each boundary is listed in \cref{tab:interfaces}. These are proposed reporting fields, not capabilities verified for every system in this survey. In particular, confidence estimates and target contact states remain design choices rather than standard VLA outputs.

\begin{table}[t]
\centering\tabsize
\caption{Proposed information to retain at stack interfaces.}
\label{tab:interfaces}
\begin{tabularx}{\columnwidth}{L{1.75cm}Y}
\toprule
Interface & Information needed for diagnosis and reuse \\
\midrule
Capture $\to$ transfer & Raw observations, hand/object state, calibration, timing and provenance \\
Transfer $\to$ learning & Action semantics, masks, embodiment metadata and feasibility tests \\
Policy $\to$ control & Target frames and units, chunk horizon, update rate and delay \\
Control $\to$ evaluation & Outcome, force, precision, damage, interventions and latency \\
Evaluation $\to$ design & Failure mode, hardware/sensor configuration, environment and recovery \\
\bottomrule
\end{tabularx}
\end{table}

\begin{figure*}[t]
\centering
\begingroup
\setlength{\tabcolsep}{7pt}
\renewcommand{\arraystretch}{1.45}
\footnotesize
\begin{tabularx}{\textwidth}{L{1.65cm}YYY}
\toprule
\textbf{Layer} & \textbf{2024} & \textbf{2025} & \textbf{2026 (through 4 November)} \\
\midrule
\rowcolor{boxgray!5}
Hardware & -- & ORCA~\cite{orcahand}; Sharpa Wave~\cite{sharpawave} & Dex5-S~\cite{unitreedex5s}; MIDAS~\cite{midas}; Aero Hand~\cite{aerohand} \\
Data & DexCap~\cite{dexcap}; Open-TeleVision~\cite{opentelevision} & AgiBot World~\cite{agibotworld}; DexWild~\cite{dexwild}; EgoDex~\cite{egodex}; DEXOP~\cite{dexop} & EgoScale~\cite{egoscale}; RealDexUMI~\cite{realdexumi}; TouchScale~\cite{touchscale} \\
\rowcolor{boxgreen!5}
Transfer & CrossDex~\cite{crossdex} & SPIDER~\cite{spider} & TactAlign~\cite{tactalign}; UniDex~\cite{unidex}; TopoRetarget~\cite{toporetarget}; OmniHOI~\cite{omnihoi} \\
Models & $\pi_0$~\cite{pi0} & Helix~\cite{figurehelix02}; GR00T N1~\cite{gr00tn1}; $\pi_{0.5}$~\cite{pi05} & DreamZero~\cite{dreamzero}; $\pi_{0.7}$~\cite{pi07}; GEN-1.5~\cite{gen15} \\
\rowcolor{boxviolet!5}
Evaluation & -- & TeleOpBench~\cite{teleopbench} & DexJoCo~\cite{dexjoco}; DexVerse~\cite{dexverse}; HandsOn~\cite{handson} \\
\bottomrule
\end{tabularx}
\endgroup
\caption{Selected milestones grouped by release year and layer. This schematic retains representative entries from the source chronology and avoids implying exact release months. A dash means no milestone was selected for that cell, not an absence of prior research. Entries are examples, not counts of research activity.}
\label{fig:timeline}
\end{figure*}

\begin{table}[t]
\centering\tabsize
\caption{Glossary of terms used in this survey.}
\label{tab:glossary}
\begin{tabularx}{\columnwidth}{L{1.9cm}Y}
\toprule
Term & Meaning \\
\midrule
Active / total DoF & Independently actuated inputs / all joints including coupled and passive ones \\
Backdrivability & Ability of external forces to move the joint through the transmission; eases compliant contact and sim-to-real \\
Retargeting & Mapping human hand motion (keypoints, joint angles) to robot joint commands \\
Embodiment gap & Mismatch in morphology, action semantics, observation, dynamics and timing between source and target \\
Robot-free capture & Data collection in which the human acts directly, with wearable hardware approximating the robot \\
Isomorphic device & Collection device with the same kinematics (and ideally sensors) as the deployed hand \\
Egocentric data & First-person video, usually with estimated or tracked 3D hand poses \\
Mid-training & Training stage on paired or aligned human--robot data between pretraining and task post-training \\
VLA & Vision-language-action model: VLM backbone plus an action decoder \\
VTLA & VLA that also consumes tactile (and often force) signals \\
World-action model & Model that predicts future observations (pixels or latents) and actions jointly \\
Contact layer (S0) & Low-level controller for tracking or contact response; tactile, force and proprioceptive inputs depend on the system \\
Progress score & Partial-credit task metric based on milestones; more lenient than binary success \\
\bottomrule
\end{tabularx}
\end{table}

\Cref{tab:glossary} defines terms used throughout.

\paragraph{L1 Hardware.} A hand is characterized by its actuation route (where the motors are and how force reaches the joints), its kinematics (number and arrangement of joints), its sensing (joint encoders, motor current, tactile skins, in-palm cameras) and its maintainability. We distinguish \emph{total} DoF (all joints, including passive or coupled ones) from \emph{active} or \emph{actuated} DoF (independently controlled inputs). Vendors mix these conventions; for example, a widely used linkage hand has 6 active and 12 total DoF~\cite{icra26hands}. Unless stated otherwise, we report active DoF.

\paragraph{L2 Data.} Data sources differ in how close they are to the deployed robot. Following the ``data pyramid'' view~\cite{datapyramid}, the top of the pyramid is robot teleoperation (directly executable, expensive), followed by wearable robot-free capture (human hands act directly, hardware approximates the robot), egocentric human video (cheapest and largest, but with the largest gap), simulation and generative data, and general vision-language data at the base.

\paragraph{L3 Cross-embodiment transfer.} We call the mismatch between a data source and the target robot the \emph{embodiment gap}. For hands it has five components: \emph{morphology} (finger count, DoF, coupling, thumb design, size); \emph{action semantics} (joint order and meaning); \emph{observation} (camera placement, appearance of the hand, tactile sensor type and layout); \emph{dynamics} (torque limits, stiffness, friction, backlash); and \emph{timing} (latency, control rate, action-chunk length). The gap arises both between humans and robots and between different robot hands.

\paragraph{L4 Learning.} Policies for dexterous hands are trained with RL (mostly in simulation), imitation learning from demonstrations, large-scale pretraining of VLA or world-action models, or combinations of these. Architectures increasingly split control into layers running at different frequencies.

\paragraph{L5 Evaluation.} Benchmarks, competitions and field deployments define what ``dexterous'' means in practice. We organize them by atomic skills (grasping, in-hand reorientation, tool use, bimanual coordination, articulated objects, deformable and fragile objects, insertion and assembly, force-controlled dosing) and by application domains.

\section{Layer 1: Hardware}
\label{sec:hardware}

\subsection{A brief lineage}
\label{sec:hw-lineage}

Today's design debates echo earlier generations of hands~\cite{piazza,controzzi}. The Robonaut~2 hand placed all 16 finger and 2 wrist actuators in the forearm and drove 12 hand DoF through braided tendons, using $n{+}1$ tendons for $n$ DoF in the primary fingers, and reached fingertip forces of about 2.25\,kg and payloads above 9\,kg~\cite{hw_robonaut2}, the template for today's forearm-actuated humanoid hands. DLR's line, from Hand~II to the Awiwi hand, was designed from the functions of the human hand rather than its anatomy~\cite{hw_dlrhands}; Awiwi's antagonistic elastic drive survived a 500\,g hammer strike and reached finger speeds of about 3{,}500\,deg/s, more than five times what its motors alone provide, and its designer argued explicitly that such robustness would allow learning by trial and error without costly damage~\cite{hw_awiwi}, an argument that \cref{sec:hw-durability} shows is more relevant than ever. EthoHand approximated the human hand with seven motors around a tendon-driven ball-joint thumb and could pass a ball between fingers while texting on a phone held in the same palm~\cite{hw_ethohand}. Underactuated prosthetic designs contributed double-stage underactuation and reconfigurable thumbs~\cite{hw_mottard17}, and the Pisa/IIT SoftHand family showed that adaptive synergies with a single actuator achieve robust grasping~\cite{hw_brlsofthand}. Soft pneumatic hands reached the full grasp taxonomy and maximal Kapandji score with RBO Hand~3, which has 16 independent degrees of actuation, an actuated palm and finger spreading~\cite{hw_rbo3}; DManus covered palm and fingertips with magnetic skin at low cost~\cite{hw_dmanus}. LEAP Hand~\cite{leaphand} and its hybrid rigid--soft successor~\cite{hw_leapv2} brought this lineage to the learning community.

\subsection{Actuation and transmission routes}
\label{sec:hw-actuation}

Where the motors sit and how they move the joints determines a hand's weight, strength, backdrivability, maintainability and, importantly for learning, how simple the mapping from motor commands to joint angles is. Six routes coexist (\cref{tab:actuation}).

\paragraph{Direct and quasi-direct drive.} Each joint has its own micro-motor and gearbox inside the finger or palm. Joints are independently controllable, backlash is low and the hand can be backdrivable, which keeps the motor-to-joint map simple and simplifies sim-to-real transfer. Recent examples include the Wuji Hand (20 active DoF, under 600\,g, a vendor-claimed endurance of more than 300k grasp cycles)~\cite{wujihand}, BrainCo Revo~3 (21 DoF, backdrivable)~\cite{revo3}, a 20-DoF fully direct-drive hand rated for a 30\,kg payload~\cite{icra26hands}, the DH-Robotics ADH-5-13 (13 DoF, modular sensors replaceable without disassembling a finger)~\cite{dhrobotics} and the Sharpa Wave (22 active DoF, 1:1 human size, a dynamic tactile array on each finger)~\cite{sharpawave}. The academic MIDAS hand (13 active of 16 joints, 283 three-axis taxels, BOM below US\$3k) ships with a simulation model and teleoperation pipeline~\cite{midas}. The main limitations are heat, limited per-finger force and the cost of 16--20 small motors.

\paragraph{Linkage.} Micro linear actuators drive coupled linkages, typically giving 6 active of 11--12 total DoF. Linkage hands are stiff, strong, durable and cheap, and dominate early humanoid deployments; their coupling limits in-hand dexterity and abduction. A six-DoF hand of this type is sold for about US\$930~\cite{icra26hands}, and the ByteDexter hand uses linkages at 20 DoF for teleoperated data collection~\cite{bytedexter}. Recent academic designs show that linkages are not inherently limited to a few actuators: DexLink drives 20 joints with 16 independent actuators at 320\,g and below US\$400 while reaching the maximum Kapandji score and all 33 grasp types~\cite{hw_dexlink}, and SyLink uses motion capture to find human joint correlations and decouples flexion from abduction with a spherical four-bar mechanism, driving 19 joints with 11 actuators at 520\,g~\cite{hw_sylink}.

\paragraph{Tendon drive.} Motors in the forearm pull tendons (often sheathed) that act on the fingers, as in the Shadow Hand~\cite{shadowhand}. This keeps the hand light and compliant and makes high DoF counts feasible, at the price of friction, creep, slack and a non-linear joint--motor map that needs tensioning and calibration. Tesla's Optimus V3 hand, according to patent analyses, moves 22 hand DoF onto 25 forearm actuators per arm and introduces a wrist routing that limits tendon-length change under combined yaw and pitch~\cite{teslav3}. Open-source tendon hands focus on maintainability: ORCA uses joints that dislocate before they break, self-calibrating tendon routing and a ratchet tensioner, and ran 10{,}000 consecutive cycles without failure~\cite{orcahand}; RUKA uses an underactuated tendon design and RUKA-v2 adds a parallel wrist~\cite{ruka,rukav2}; MM-Hand moves the motors to a remote housing with quick-release tendon connectors and analyzes sheath friction~\cite{mmhand}; CRAFT combines soft joints with rigid links for under US\$600~\cite{crafthand}; and a 17-DoF antagonistic design drives both flexion and extension without return springs~\cite{antagonistichand}. The tendon-to-joint map is increasingly handled with learning and proprioception rather than hand-tuned friction models: a transformer that predicts tendon forces within 3\% of maximum motor force reduces the sim-to-real gap by 41\% and improves fingertip tracking of an RL controller by 50\% on real tendon-driven fingers~\cite{hw_tendonforce}; series-elastic tendons estimate contact timing, joint angles and object stiffness without vision or touch~\cite{hw_tendonprop}; model-based estimation recovers joint positions from tendon displacement and tension~\cite{hw_tendonmodel}; and the DexHand~021 (12 active and 7 passive DoF, 1\,kg) uses proprioceptive compliance control with force estimation errors below 0.2\,N~\cite{hw_dexhand021}. Artificial muscles are another remote-actuation option: electrohydraulic Peano-HASEL actuators in the forearm drive an inherently compliant musculoskeletal hand that detects grasps from operating current alone~\cite{hw_hasel}.

\paragraph{Hybrid drive.} 2026 product launches show a clear shift to hybrids: tendons for high-force flexion and in-palm direct drive or linkages for abduction and opposition. Xiaomi's 21-DoF hand uses 26 motors with tendon-tension control and was designed from an analysis of 1{,}827 factory standard operating procedures~\cite{xiaomihand}. Research hands explore other hybrids, such as shape-memory-alloy extension combined with motor-driven flexion~\cite{cyjhand}, or a split of five forearm motors for grip force and ten palm motors for fine manipulation in a 15-DoF, 1.4\,kg hand~\cite{hw_15dof}.

\paragraph{Hydraulic and servo-based.} Sanctuary AI's 21-DoF hand uses miniature hydraulic valves and was shown performing zero-shot in-hand cube reorientation using only the fingertips~\cite{sanctuaryhand}; in academia, DexCo uses soft hydraulic actuation with hydrostatic compliance modeling and reports 34.4\,N repeatable fingertip force and 0.03\,mm repeatability~\cite{hw_dexco}. At the other end, servo-based open hands such as LEAP~\cite{leaphand}, Aero Hand Open~\cite{aerohand} and AmazingHand~\cite{amazinghand} trade anthropomorphic proportions and strength for very low cost and easy repair.

\begin{table*}[t]
\centering\tabsize
\caption{Actuation and transmission routes for dexterous hands. Product examples are illustrative; specifications are vendor-reported.}
\label{tab:actuation}
\begin{tabularx}{\textwidth}{L{2.0cm}L{3.3cm}L{3.6cm}YYL{2.2cm}}
\toprule
Route & Principle & Examples & Strengths & Weaknesses & Typical use \\
\midrule
Direct / quasi-direct & One micro-motor + reducer per joint, in finger or palm & Wuji Hand~\cite{wujihand}, Revo~3~\cite{revo3}, Sharpa Wave~\cite{sharpawave}, ADH-5-13~\cite{dhrobotics}, MIDAS~\cite{midas}, Allegro~\cite{allegro} & Independent joints, low backlash, backdrivable, simple joint--motor map; good for RL and sim-to-real & Limited space and heat dissipation; many motors; modest finger force & Research, RL, data-collection platforms \\
Linkage & Linear actuators drive coupled linkages (e.g., 6 active / 12 total) & Linkage hands on early humanoids~\cite{icra26hands}, ByteDexter~\cite{bytedexter} & Stiff, strong, durable, cheap, easy to mass-produce & Coupled joints, weak abduction, limited in-hand dexterity & Industrial picking, logistics \\
Tendon & Forearm motors pull (sheathed) tendons & Shadow~\cite{shadowhand}, Optimus V3~\cite{teslav3}, ORCA~\cite{orcahand}, RUKA~\cite{ruka,rukav2}, MM-Hand~\cite{mmhand}, CRAFT~\cite{crafthand} & Light, low inertia, compliant, high DoF & Friction, creep, slack; tensioning and calibration; non-linear mapping & Anthropomorphic and humanoid hands \\
Hybrid & Tendons for flexion; direct drive/linkage for abduction and opposition & Xiaomi 21-DoF~\cite{xiaomihand}, CYJ Hand-0~\cite{cyjhand} & Combines force, low weight and precision per joint & Highest system complexity and maintenance cost & 2026 high-end products \\
Hydraulic & Miniature hydraulic valves & Sanctuary AI~\cite{sanctuaryhand} & High power density, fine force resolution & Complexity, cost, leakage risk & Research platforms \\
Servo (open-source) & Off-the-shelf bus servos at the joints & LEAP~\cite{leaphand}, Aero Hand Open~\cite{aerohand}, AmazingHand~\cite{amazinghand} & Very cheap, easy to build and repair & Bulky, non-anthropomorphic proportions, limited force and life & Entry-level research, teaching \\
\bottomrule
\end{tabularx}
\end{table*}

\begin{table*}[t]
\centering\tabsize
\caption{Specifications of representative commercial and humanoid-integrated dexterous hands (vendor-reported or press-reported; ``--'' = not disclosed). DD: direct/quasi-direct drive; Ten.: tendon; Link.: linkage; Hyb.: hybrid; Hyd.: hydraulic.}
\label{tab:commercialhands}
\begin{tabularx}{\textwidth}{L{2.55cm}L{0.75cm}L{0.6cm}L{1.9cm}L{1.15cm}L{1.9cm}L{3.0cm}Y}
\toprule
Hand (maker) & Year & Route & DoF & Mass & Force / payload & Sensing & Notes \\
\midrule
RH56 series (Inspire)~\cite{mir2026,yicaireliability} & 2023+ & Link. & 6 active / 12 & 0.5--0.6\,kg & -- & Optional fingertip touch & Standard hand on early humanoids; $>$10k units shipped in 2025 (reported) \\
RH524J1 (Inspire)~\cite{inspire524} & 2026 & Hyb. & 24 (22 hand + 2 wrist) & -- & -- & Proximity, touch, depth & Tendon-hybrid; modular, thermal management \\
O6 (Linkerbot)~\cite{linkerprice,icra26hands} & 2025 & Link. & 6 & -- & $>$3\,kg per finger & -- & Volume product, about US\$930 \\
O30 (Linkerbot)~\cite{icra26hands,waic26hands} & 2026 & DD & 20 & 740\,g & 30\,kg rated & -- & Each joint independently driven; aimed at RL \\
OmniHand 3 Ultra-M (AGILINK)~\cite{agilinkduet,waic26hands} & 2026 & DD & 20 & 630\,g & -- & 5 fingertip visuo-tactile + 300+ palm taxels & Paired with DUET contact layer \\
Revo 3 (BrainCo)~\cite{revo3} & 2026 & DD & 21 & -- & 70\,N grip & Full-palm + fingertip visuo-tactile & Backdrivable; 33 grasp types \\
Wuji Hand / Hand 2~\cite{wujihand} & 2025--26 & DD & 20 (4 per finger) & $<$600\,g & $\sim$15\,N fingertip & -- & 300k+ cycles; about US\$16k \\
DexHand021 Pro (DexRobot)~\cite{dexrobot} & 2026 & Ten. & 22 & -- & 50\,N & Full-palm multimodal & Bidirectional tendons; 300k+ cycles \\
Dex5-S (Unitree)~\cite{unitreedex5s} & 2026 & DD & 22 & 620\,g & -- & Touch (Pro version) & Human-sized; about US\$6.5k \\
21-DoF hand (Xiaomi)~\cite{xiaomihand} & 2026 & Hyb. & 21 (26 motors) & -- & -- & 1{,}500 taxels & Designed from 1{,}827 factory SOPs \\
Optimus V3 hand (Tesla)~\cite{teslav3,teslav3actuators} & 2026 & Ten. & 22 + wrist & -- & -- & Fingertip touch & 25 forearm actuators per arm \\
Figure 03 hand~\cite{figure03} & 2025 & Ten. & $\sim$20 & -- & -- & Palm camera, 3\,g fingertip force & Soft fingertips \\
1X hand~\cite{onexhand} & 2026 & Ten. & 25 (incl.\ wrist) & -- & -- & -- & Modeled on human tendons \\
Phoenix hand (Sanctuary)~\cite{sanctuaryhand} & 2025--26 & Hyd. & 21 & -- & -- & Barometric, 5\,mN & Fingertip-only reorientation demo \\
Sharpa Wave / W02~\cite{sharpawave,sharpaw02} & 2025--26 & DD & 22 / 21 & W02 $<$750\,g & 20\,N fingertip & Dynamic tactile array per finger; full-hand (W02) & 1:1 human size; GR00T reference humanoid \\
ADH-5-13 (DH-Robotics)~\cite{dhrobotics} & 2026 & DD & 13 & 735\,g & -- & Fingertip touch, expandable & Sensors swappable without disassembly \\
DexH13 (PaXini)~\cite{paxinimall,paxinipxcap} & 2025 & DD & 13 & -- & -- & 1{,}140 Hall-effect cells & Paired with data-collection glove \\
Shadow Dexterous Hand~\cite{shadowhand} & classic & Ten. & 20 active / 24 & $\sim$4.3\,kg (with forearm) & -- & Various options & $>$US\$100k; research benchmark \\
Allegro Hand V5~\cite{allegro} & classic & DD & 16 (4 fingers) & $\sim$1\,kg & -- & Optional fingertip touch & Common RL platform; about US\$17k \\
Ability Hand (PSYONIC)~\cite{hato} & classic & Link. & 6 & $\sim$0.5\,kg & -- & Fingertip touch & Prosthetic origin \\
ZWHAND~\cite{zwhand} & 2025 & Link. & 17--18 & $\sim$0.85\,kg & 6\,kg & E-skin & About US\$12.5k \\
\bottomrule
\end{tabularx}
\end{table*}

\begin{table*}[t]
\centering\tabsize
\caption{Academic and open-source dexterous hands (2023--2026). Cost is reported BOM or kit price.}
\label{tab:openhands}
\begin{tabularx}{\textwidth}{L{2.3cm}L{0.9cm}L{1.6cm}L{1.6cm}L{2.2cm}Y}
\toprule
Hand & Year & Route & DoF & Mass / cost & Highlights \\
\midrule
LEAP Hand~\cite{leaphand} & 2023 & Servo DD & 16 (4 fingers) & 600\,g / $\sim$US\$2k & Universal MCP abduction; 4-h assembly; $\sim$1/8 Allegro cost \\
ORCA~\cite{orcahand,orcacommercial} & 2025 & Tendon & 17 & -- / $<$CHF\,2k & Dislocating joints, self-calibration; 10k cycles; commercial lite/standard/touch \\
RUKA / RUKA-v2~\cite{ruka,rukav2} & 2025--26 & Tendon & 15 underact.\ (v1); + 2-DoF wrist (v2) & $<$US\$1.3k--1.5k & Fully open CAD; learned controllers \\
CRAFT~\cite{crafthand} & 2026 & Tendon & 15 actuated & $<$US\$600 & Hybrid hard--soft compliance; 33/33 GRASP types \\
MM-Hand~\cite{mmhand} & 2026 & Remote tendon & 21 & -- & Quick-release tendons; in-palm stereo; sheath-friction analysis \\
MIDAS~\cite{midas} & 2026 & DD & 13 of 16 & 700\,g / $<$US\$3k & 283 three-axis taxels; sim model, retargeting, teleop pipeline \\
Aero Hand Open~\cite{aerohand} & 2026 & Tendon (7 servos) & 16 joints & 374\,g / US\$314 & Simulation-ready; fully 3D printed \\
Antagonistic hand~\cite{antagonistichand} & 2026 & Antagonistic tendon & 17 & 220\,g & No return springs; palmar and dorsal grasps; 33 grasp types \\
Tactile tendon hand~\cite{tactiletendonhand} & 2026 & Tendon & 5 fingers & -- & Distributed bimodal touch on phalanges \\
CYJ Hand-0~\cite{cyjhand} & 2025 & SMA + motor & 21 & -- & SMA extension, motor flexion; printed metal skeleton \\
ByteDexter~\cite{bytedexter} & 2025 & Linkage & 20 & -- & Retargeted teleoperation for long-horizon data \\
RAPID Hand~\cite{rapidhand,hw_rapidproto} & 2025 & -- & 20 & -- & Low-cost 20-DoA prototype; co-designed hand, perception and teleoperation \\
DexLink~\cite{hw_dexlink} & 2026 & Linkage & 16 actuators / 20 joints & 320\,g / $<$US\$400 & Max.\ Kapandji; 33/33 grasp types \\
SyLink~\cite{hw_sylink} & 2026 & Linkage & 11 actuators / 19 joints & 520\,g / $\sim$US\$400 & Mocap-derived synergy coupling \\
BiDexHand~\cite{hw_bidexhand} & 2025 & Tendon & 16 + 5 coupled & -- & 33/33 grasps; 9/11 Kapandji; 2.14\,N fingertip \\
ISyHand~\cite{hw_isyhand} & 2025 & Servo & articulated palm & $\sim$US\$1.3k & Palm beats fixed-palm variant in RL \\
DexHand 021~\cite{hw_dexhand021} & 2025 & Tendon & 12 active + 7 passive & 1\,kg & Proprioceptive compliance; $<$0.2\,N force error \\
AmazingHand~\cite{amazinghand} & 2025 & Servo & 8 (4 fingers) & 400\,g / $\sim$US\$231 & Ultra-low-cost entry hand \\
\bottomrule
\end{tabularx}
\end{table*}

Specifications of representative commercial and open-source hands are collected in \cref{tab:commercialhands,tab:openhands}.

\subsection{How many degrees of freedom?}
\label{sec:hw-dof}

Commercial hands are marketed with anywhere from 6 to 42 DoF, but flagship designs are converging on roughly 20--22 active DoF (Sharpa Wave 22~\cite{sharpawave}, Dex5-S 22~\cite{unitreedex5s}, Optimus V3 22~\cite{teslav3}, Revo~3 21~\cite{revo3}, Xiaomi 21~\cite{xiaomihand}). The most explicit evidence comes from Xiaomi's industrial study, which reports that across 1{,}827 factory procedures, hands with 17--21 DoF achieved 15--38\% higher average stable-grasp success than 6--15-DoF hands, with diminishing returns beyond 23 DoF, and that joints with correlation above 0.8 (e.g., the distal interphalangeal joint following roughly one third of the proximal one) are good candidates for underactuated coupling~\cite{xiaomihand}. This is a vendor study rather than a controlled academic experiment, but it is consistent with the grasp coverage results of open hands: CRAFT and an antagonistic 17-DoF hand cover all 33 grasp types of the GRASP taxonomy~\cite{feix,crafthand,antagonistichand}. Grasp-taxonomy coverage, however, no longer discriminates between designs: low-cost tendon and linkage hands now routinely complete all 33 grasp types~\cite{hw_bidexhand,hw_dexlink,crafthand}. More informative are fingertip force (a 16-DoF open tendon hand completes every grasp type yet averages only 2.14\,N peak fingertip force~\cite{hw_bidexhand}), opposition range and durability.

The \emph{thumb} remains the hardest joint. BiDexHand misses two of eleven Kapandji poses because of limited carpometacarpal range~\cite{hw_bidexhand}; a soft biomimetic saddle-joint thumb scores 9/10~\cite{hw_anthrothumb}; and design methods place the thumb by workspace analysis~\cite{hw_cerruti15} or make it reconfigurable~\cite{hw_mottard17}. Optimizing phalanx length ratios against manipulability and workspace metrics shows that phalanges do not contribute equally to dexterity~\cite{hw_phalanx}. The \emph{palm} and \emph{wrist} are becoming design variables: ISyHand's articulated palm significantly outperforms a fixed-palm variant of the same hand in RL cube reorientation~\cite{hw_isyhand}, and wrists are integrated into synergy hands~\cite{hw_modelw} or added to low-cost tendon hands with quantified ranges~\cite{hw_crafwrist,rukav2}. Whether anthropomorphism itself is necessary is contested: a review of 125 papers from 2019--2025 concludes that in-hand manipulation does not benefit from anthropomorphic design and that mechanism complexity correlates with task breadth~\cite{hw_needanthro}, whereas musculoskeletal designs argue that anatomical priors help identify which structures matter for control~\cite{hw_mcr}. For learning from human data, the relevant question is narrower: how closely the hand's kinematics and camera placement must match a human's for human data to transfer (\cref{sec:data-scaling}).

We expect two tiers to persist: 6--12-DoF hands for high-throughput industrial grasping and 20--22-DoF hands for research, data collection and general-purpose humanoids.

A related design question is \emph{where the actuators live}. Humanoid makers increasingly move large motors into the forearm, as in Optimus V3~\cite{teslav3}, yielding an integrated arm--hand module with low distal inertia. Hands sold as standalone products, by contrast, mostly keep actuation in the palm so that they can be mounted on arbitrary arms.

\subsection{Soft, compliant and variable-stiffness hands}
\label{sec:hw-soft}

Compliance protects hands and objects and simplifies teleoperation and learning. Beyond RBO Hand~3~\cite{hw_rbo3}, recent examples include the Tactile SoftHand-A, a 3D-printed underactuated hand with antagonistic tendons to open and close actively and an integrated vision-based tactile sensor~\cite{hw_tactilesofthand}; a LEGO-based educational synergy hand at the lowest end of the cost spectrum~\cite{hw_edusofthand}; pneumatic hands with single-step printed exoskeletons~\cite{hw_multimaterial}; and soft--rigid hybrid fingers whose stiffness is set by control gains rather than hardware, which completed 28 of 33 grasp postures and supported diffusion-policy learning from glove teleoperation~\cite{hw_norton}. Variable-stiffness joints based on beaded-string jamming span 0.48--1.95\,Nm/rad and a gripper integrating the joint reached 60\% higher peg-in-hole success than a rigid one~\cite{hw_jamming}. Co-design is moving from simulation to physical validation: a genetic algorithm with policy transfer evaluated nearly 400 soft-hand designs and validated them with more than 900 teleoperated tasks~\cite{hw_mannam}, and multi-parameter benchmarking of candidate fingers (joint, bone, skin and sensor placement) precedes assembling a full hand~\cite{hw_modularfinger}.

\subsection{Tactile sensing}
\label{sec:hw-tactile}

Touch moved from an option to a selling point in 2026. Products advertise full-palm coverage: 1{,}500 taxels on Xiaomi's hand~\cite{xiaomihand}, 1{,}140 Hall-effect cells on the PaXini DexH13~\cite{paxinimall}, fingertip visuo-tactile sensors plus more than 300 palm taxels on the AGILINK OmniHand~3~\cite{agilinkduet}, and a dynamic tactile array per finger on the Sharpa Wave, with a full-hand tactile successor~\cite{sharpawave,sharpaw02}. Sanctuary reports a barometric tactile sensitivity of 5\,mN~\cite{sanctuaryhand}, and Figure~03 reports fingertip force sensing at the 3\,g level~\cite{figure03}. Three sensing principles dominate: \emph{piezoresistive} skins (cheap, fast and simple, and the mainstream choice for mass production), \emph{capacitive} and \emph{magnetic/Hall} arrays, and \emph{visuo-tactile} sensors that image the deformation of an elastomer with an internal camera (highest spatial resolution, used for fragile objects and slip). Academic hands increasingly integrate distributed sensing at the phalanges~\cite{tactiletendonhand} and full-hand arrays~\cite{midas}. Full coverage is not always necessary: task-driven pruning reduced a Shadow Hand array from 92 to 21 sensors while retaining 93\% of performance, with thumb, ring and little-finger sensors contributing most~\cite{hw_tactileopt}. The fragmentation of sensor types is a major obstacle to reusing tactile data across hands; we discuss sensors, simulation and tactile learning in depth in \cref{sec:touch}.

\subsection{Open-source and academic hands}
\label{sec:hw-open}

Open-source hands have evolved from ``it moves'' to ``it is ready for learning''. LEAP (2023) introduced a universal metacarpophalangeal abduction mechanism, about US\$2k and a four-hour assembly, and outperformed the Allegro Hand at roughly one eighth of its cost~\cite{leaphand}. Since 2025, designs ship with what a learning pipeline needs: simulation models (Aero Hand Open is released ``simulation-ready'' at US\$314~\cite{aerohand}; MIDAS includes control and tactile APIs, retargeting and teleoperation~\cite{midas}), integrated touch (ORCA's tactile variant~\cite{orcahand,orcacommercial}), wrists (RUKA-v2~\cite{rukav2}), in-hand stereo cameras (MM-Hand~\cite{mmhand}) and maintainability features. Some designs are being commercialized: ORCA offers 9-, 17-DoF and tactile versions at US\$1.5k--6.1k~\cite{orcacommercial}. Hands are also being co-designed with data collection and with policies: RAPID Hand co-optimizes hand, perception and teleoperation interface~\cite{rapidhand}, and House of Dextra pretrains a morphology-conditioned policy over many generated hand shapes to evaluate candidate designs~\cite{houseofdextra}.

\subsection{Durability, cost and supply chain}
\label{sec:hw-durability}

\paragraph{Durability is the first deployment bottleneck.} Integrators report that some popular hands fail after about 50 grasps of a 5\,kg object; a warehouse robot performs 8{,}000--10{,}000 grasps per day while leading products are rated for 300k--500k cycles, which suggests a potential mismatch with continuous duty, but cannot by itself establish replacement intervals without matched loads and failure definitions~\cite{yicaireliability}. Figure reported that the forearm was the most common failure point in its BMW pilot~\cite{figure03}. Vendor claims of 300k cycles~\cite{wujihand,dexrobot} sit near today's passing line; industrial deployment will require million-cycle lifetimes, load-specified test conditions and modular quick-swap designs.

\begin{figure*}[t]
\centering
\includegraphics[width=\textwidth]{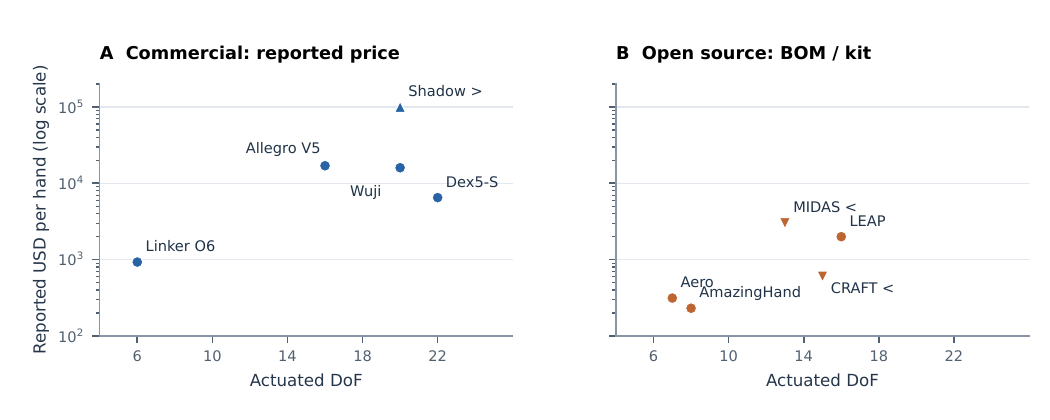}
\caption{Reported cost and actuated DoF for selected hands, with commercial prices separated from open-source BOM/kit costs~\cite{linkerprice,unitreedex5s,wujihand,allegro,shadowhand,amazinghand,aerohand,crafthand,leaphand,midas}. Approximate or upper/lower-bound source figures are plotted at their stated reference values; Shadow is reported above US\$100k, CRAFT below US\$600 and MIDAS below US\$3k. Costs differ in included components, support, assembly and date, so these are not equivalent procurement quotes. The cross-sectional plot does not establish a temporal price decline or a performance frontier.}
\label{fig:pricedof}
\end{figure*}

\begin{figure}[t]
\centering
\includegraphics[width=\columnwidth]{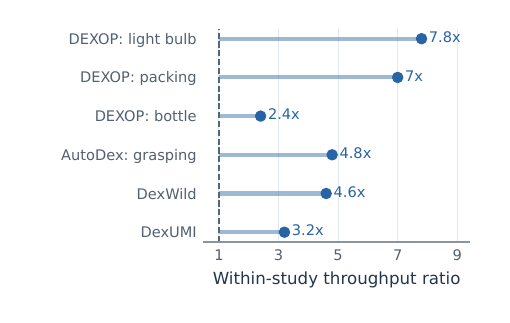}
\caption{Within-study collection-throughput ratios~\cite{dexop,autodex,dexwild,dexumi}. Rows are grouped by source, not ranked. Each ratio uses its own teleoperation baseline (dashed line at one); tasks, acceptance criteria and operators differ. DexDirect is excluded because its reported quantity is the number of successful demonstrations, not an explicitly normalized throughput.}
\label{fig:throughput}
\end{figure}

\paragraph{Cost.} A complete low-cost bimanual research platform illustrates where money goes: about US\$5k for two robot hands, US\$7k for the teleoperation system and US\$16k for two arms~\cite{dep_bidex}. \Cref{fig:pricedof} plots reported prices against actuated DoF. Reported prices span a wide range across products: established research hands are quoted around US\$16k--100k or more~\cite{shadowhand,allegro,wujihand}, while the 22-DoF Dex5-S is reported at about US\$6.5k~\cite{unitreedex5s}, while open-source BOMs reach a few hundred dollars~\cite{aerohand,amazinghand,crafthand}. The dexterous hand is still among the most expensive single components of a humanoid. Its cost is dominated by 16--20 coreless micro-motors (6--8\,mm diameter) per high-DoF hand, a market led by a few European suppliers; domestic Chinese alternatives are reported at 30--40\% of import prices but still trail at the smallest sizes~\cite{bom36kr}.

\paragraph{Supply chain.} Beyond motors, transmission components (micro lead screws and planetary roller screws for linear drives, tendons and sheaths) and sensing components (six-axis force sensors, electronic skins, joint encoders) form the rest of the bill of materials, and hand makers increasingly develop actuators in-house to control cost~\cite{bom36kr,mir2026}. Industry analyses describe an ``impossible triangle'' among dexterity, payload and cost~\cite{impossibletriangle}, a market that is fragmented with small revenues per vendor, and valuations that run ahead of revenue~\cite{linkervaluation}. Expert opinion converges on roughly 20 DoF as necessary for human-level dexterity, with six-DoF hands not strictly dexterous and 40+ DoF largely redundant~\cite{dofopinion}. Tactile sensors attracted dozens of financing rounds in 2025--2026~\cite{tactilefunding}.

\paragraph{Volume.} Market trackers estimate 19.2k hands sold in China in 2025, with forecasts of 70.2k in 2026 and 430k by 2030~\cite{ggii}; individual vendors report cumulative shipments in the 10k range~\cite{mir2026}. These figures are not audited and should be read as indicators of momentum rather than market size.

\takeaway{Hardware is converging on $\sim$20--22 active DoF with hybrid actuation and full-hand touch for high-end hands, while 6--12-DoF linkage hands serve industrial picking. Cost is falling fast; durability under realistic loads, not DoF, is now the binding constraint. For learning, backdrivable direct-drive joints and human-like proportions and camera placement make sim-to-real and human-data reuse easier (\cref{sec:data-scaling}).}

\section{Layer 2: Data}
\label{sec:data}

Data for dexterous manipulation can be organized by how close the recorded actions and observations are to those of the deployed robot. Closer sources are more directly usable but more expensive per hour; more distant sources scale better but require bridging an embodiment gap (\cref{sec:crossemb}). \Cref{tab:datasystems} summarizes representative systems, \cref{fig:datascale} the reported data scales, and \cref{tab:datacompare} a qualitative comparison of routes.

\begin{table*}[t]
\centering\tabsize
\caption{Representative dexterous data-collection systems and datasets (2024--2026). Scale as reported; ``h'' = hours, ``demos'' = demonstrations.}
\label{tab:datasystems}
\begin{tabularx}{\textwidth}{L{2.35cm}L{0.95cm}L{1.55cm}L{4.6cm}YL{2.2cm}}
\toprule
System & Date & Route & Hardware & Key idea / finding & Scale \\
\midrule
DexCap~\cite{dexcap} & 2024-03 & Glove & Mocap glove, chest RGB-D + SLAM & Portable robot-free capture with human-in-the-loop correction & Tens of minutes/task \\
Open-TeleVision~\cite{opentelevision} & 2024-07 & VR teleop & Vision Pro, active stereo neck & Immersive active vision & Tens of demos/task \\
Bunny-VisionPro~\cite{bunnyvisionpro} & 2024-07 & VR teleop & Vision Pro, vibrotactile & Haptic cues and safety constraints & -- \\
ACE~\cite{ace} & 2024-08 & Exoskeleton & Arm exo + hand tracking & Cross-platform visual exoskeleton & -- \\
DOGlove~\cite{doglove} & 2025-02 & Glove & Open force-feedback glove & Manipulation under occlusion via haptics & -- \\
AgiBot World~\cite{agibotworld} & 2025-03 & Teleop factory & 100 robots, VR teleop & Standardized pipeline; scaling of GO-1 & 1M+ traj., 2{,}976\,h \\
DexWild~\cite{dexwild} & 2025-05 & Glove + palm cams & EM glove, mirrored palm cameras & 4.6$\times$ teleop speed; 1:2 co-training & 9{,}290 demos, 93 envs \\
EgoDex~\cite{egodex} & 2025-05 & Egocentric & Vision Pro tracking & Native 3D finger tracking & 829\,h, 194 tasks \\
DexUMI~\cite{dexumi} & 2025-05 & Exoskeleton & Hand-specific exo, wrist cam & Kinematic constraint + inpainting & Hundreds/task \\
DEXOP~\cite{dexop} & 2025-09 & Passive exo + hand & Exo linked to passive robot hand & Perioperation; 2.4--8$\times$ teleop & -- \\
MILE~\cite{mile} & 2025-11 & Isomorphic exo & Exo + LEAP-style hand, visuo-tactile & $<$1$^\circ$ joint error; touch +25\% & Multimodal dataset \\
WIYH~\cite{wiyh} & 2025-12 & Wearable & Vision + mocap + touch suite & Auto-annotation, tactile VLTA data & 1{,}000+\,h \\
UniHand-2.0~\cite{beingh05} & 2026-01 & Human + robot & Egocentric + 30 robots & Human hand as a ``native'' action space & 35k\,h \\
EgoScale~\cite{egoscale} & 2026-02 & Egocentric & Head cam; glove + tracker for aligned set & Log-linear scaling; +54\% & 20{,}854\,h \\
DexViTac~\cite{dexvitac} & 2026-03 & Glove + touch & Fisheye, glove, tactile, T265 & 248+ demos/h; $>$85\% success & 2{,}400+ demos \\
XRZero-G0~\cite{xrzero} & 2026-04 & Wearable & VR tracking, backpack compute & Closed loop, 85\% valid data & 2{,}000+\,h \\
DEX-Mouse~\cite{dexmouse} & 2026-04 & Handheld & Forearm-mounted hand, $<$US\$150 & Calibration-free, kinesthetic feedback & Open BOM \\
RealDexUMI~\cite{realdexumi} & 2026-06 & Deployed hand & 11-DoF hand, in-hand cam, touch & Zero gap; 88.75\% (70\% w/o touch) & 100+\,h \\
YUBI~\cite{yubi} & 2026-06 & UMI & Finger-aligned two-finger gripper & Largest UMI-style bimanual set & 8{,}434\,h \\
AutoDex~\cite{autodex} & 2026-06 & Automated & 20 RGB cams, arm + hand & Unattended loop, 4.8$\times$ teleop & 3{,}593 grasps \\
DexDirect~\cite{dexdirect} & 2026-07 & Kinesthetic + vision & Gravity-comp. arm, webcam fingers & 3.2$\times$ TeleDex, 17.2$\times$ AnyTeleop & 481 demos (study) \\
AdvDex~\cite{advdex} & 2026-08 & Glove + touch & 29 encoders, Hall arrays & Joint-aligned actions, adversarial & 100k+ traj. \\
DITTO / -X~\cite{ditto,dittox} & 2026-09 & Motorized exo & 1:1 exo for 7-DoF hand & Bilateral force feedback; DAgger & -- \\
DexPort~\cite{dexport} & 2026-10 & Deployed hand & Handheld grip with deployment hand & Native command/state recording & -- \\
TouchScale~\cite{touchscale} & 2026-10 & Tactile glove & 880 taxels/hand + ego + wrist RGB & Visuo-tactile mid-training & 500\,h \\
\midrule
Sunday~\cite{sundayact1} & 2025-11 & Industry glove & US\$200 glove matching robot hand & Zero robot data for ACT-1 & $\sim$10M episodes~\cite{sunday36kr} \\
Generalist~\cite{gen15,generalistfunding} & 2025--26 & Industry wearable & Pincer-style data hands & Pretraining mainly on human demos & 270k$\to$500k+\,h~\cite{generalistprofile} \\
Liber-0~\cite{liber0} & 2026 & Industry glove & 131\,g glove, ego + 2 wrist cams & Glove $>$ 200$\times$ web video & 2.2M\,h total (mostly web) \\
\bottomrule
\end{tabularx}
\end{table*}

\begin{table}[t]
\centering\tabsize
\caption{Qualitative comparison of data routes. ``+'' favorable, ``$-$'' unfavorable.}
\label{tab:datacompare}
\begin{tabularx}{\columnwidth}{L{1.55cm}YYYY}
\toprule
 & Teleop & Wearable & Egocentric & Sim/gen. \\
\midrule
Cost per hour & $-$ & + & ++ & ++ (marginal) \\
Throughput & $-$ & + & ++ & ++ \\
Executability & ++ & + (iso.) / $\pm$ & $-$ & $\pm$ \\
Embodiment gap & none & small--med. & largest & sim2real \\
Contact / force & $-$ & ++ & $-$ & $\pm$ \\
Scene diversity & $-$ & + & ++ & $\pm$ \\
Typical role & post-train, eval & mid-train & pretrain & amplify \\
\bottomrule
\end{tabularx}
\end{table}

\subsection{Teleoperation}
\label{sec:data-teleop}

In teleoperation a human controls the robot and the robot's own observations and joint commands are recorded, which gives directly executable data with no embodiment gap. Interfaces fall into three groups.

\paragraph{Vision- and VR-based hand tracking.} Headsets such as Apple Vision Pro and Meta Quest provide stereo video and hand keypoints, which are retargeted to robot joints~\cite{dexpilot,anyteleop}. Representative systems include OPEN TEACH~\cite{openteach}, Open-TeleVision with head-following active stereo~\cite{opentelevision}, Bunny-VisionPro with vibrotactile feedback and collision constraints~\cite{bunnyvisionpro}, HATO for bimanual visuotactile teleoperation with repurposed prosthetic hands~\cite{hato}, monocular whole-body shadowing~\cite{humanplus,omnih2o}, and phone-based interfaces~\cite{teledex}. These systems are cheap and fast to deploy, but finger tracking jitters under occlusion and there is no force feedback, so contact-rich tasks suffer: in the DexDirect user study, a vision-based baseline achieved zero success on whiteboard wiping and typing, and DexDirect collected 3.2$\times$ more successful demonstrations than TeleDex and 17.2$\times$ more than AnyTeleop~\cite{dexdirect}.

\paragraph{Data gloves.} Commercial electromagnetic gloves track 25 skeletal joints at 120\,Hz without optical occlusion and have become a de facto standard in industry~\cite{manusicra}; they are, however, sensitive to nearby metal and motors. Research gloves add force or tactile feedback at low cost: DOGlove~\cite{doglove}, CDF-Glove~\cite{cdfglove}, the tactile-feedback-array glove TAG~\cite{feelrobot}, and Glovity, which renders wrist wrenches~\cite{glovity}. Wrist pose is tracked separately with trackers, arm exoskeletons or SLAM cameras.

\paragraph{Exoskeletons and isomorphic leader devices.} Exoskeletons give joint-level, low-latency mappings and can provide force feedback, at the cost of being tied to one target hand. The lineage runs from ACE~\cite{ace}, HOMIE~\cite{homie}, NuExo~\cite{nuexo} and Exo-ViHa~\cite{exoviha} to MILE, which co-designs a mechanically isomorphic exoskeleton and robot hand with sub-degree joint error and shows that fingertip visuo-tactile sensing raises success by 25\% on average~\cite{mile}. DITTO maps a motorized exoskeleton one-to-one onto a kinematically equivalent 7-DoF hand, unifying handheld collection and bilateral teleoperation with joint force feedback~\cite{ditto}; DITTO-X adds reverse teleoperation for DAgger-style interventions~\cite{dittox}. DexterityGen takes a complementary route: a pretrained RL controller executes low-level finger motion while the operator only provides high-level intent~\cite{dexteritygen}.

\paragraph{Lineage and leader devices.} Hand teleoperation for learning goes back to mixed-reality headset tracking with non-parametric imitation (Holo-Dex, six dexterous tasks learned within hours)~\cite{dep_holodex}, single-RGB-camera imitators trained on internet hand video (Robotic Telekinesis)~\cite{dep_telekinesis}, and DIME, which sidesteps unstable retargeting by fitting a per-user simulated hand to the operator and converting trajectories to different real hands afterwards~\cite{dep_dime}. For arms, kinematically matched leader devices such as GELLO proved more reliable and efficient than VR controllers or space mice~\cite{dep_gello}, and wearable arm exoskeletons showed that 3 minutes of teleoperation plus in-the-wild data can match more than 20 minutes of teleoperation alone~\cite{dep_airexo}; AirExo-2 adds visual and action adaptors that turn in-the-wild exoskeleton data into pseudo-robot demonstrations with 1.7\,mm action error, on a US\$600 platform~\cite{dep_airexo2}. Bidex extends ALOHA-style leader--follower setups to two dexterous hands on arms (more than 50 DoF in total) with mocap gloves and teacher arms, collecting better data faster than headset-based systems~\cite{dep_bidex}; modular mobile teleoperation interfaces~\cite{dep_telemoma} and whole-body humanoid systems with learned, tuning-free hand retargeting~\cite{dep_humdex} broaden coverage. Kinesthetic hand-over-hand teaching is another route: KineDex transfers the operator's finger motion directly to the hand with tactile signals, collects data more than twice as fast as teleoperation on two tasks and reaches 74.4\% success on contact-rich tasks with force-controlled deployment~\cite{dep_kinedex}.

\paragraph{Large teleoperated corpora.} Teleoperation also underpins large multi-embodiment datasets such as RoboMIND~\cite{robomind} and AgiBot World (more than one million trajectories and 2{,}976 hours from 100 robots, some with six-DoF hands)~\cite{agibotworld}, and initial versions of industrial VLAs~\cite{figurehelix02}. The common bottleneck is that throughput equals the number of robots times operating hours, collection is confined to labs or data factories, and operators do not touch the objects, so demonstrations lack human contact strategies.

\subsection{Retargeting, haptics and shared autonomy}
\label{sec:data-retarget}

\paragraph{Retargeting algorithms.} Retargeting maps operator hand motion to robot joints in real time and is a hidden source of demonstration quality. A systematic ablation of the factors in retargeting objectives, together with a proposed combined objective, concludes that full reproduction of human motion is impossible because of mechanical differences~\cite{dep_retargetobj}. Recent algorithms add structure: Kilohertz-Safe formulates retargeting as a convex quadratic program in joint-velocity space with linearized joint-limit and collision constraints and control barrier functions, achieving 9.05\,ms average latency with more than 95\% of frames meeting safety criteria on a 20-DoF hand~\cite{dep_kilohertz}; ResPilot learns a residual Gaussian-process correction on top of keypoint optimization from 24 calibration poses (about 4.5 minutes per operator), enlarging the reachable fingertip workspace enough to enable finger gaiting~\cite{dep_respilot}; EFOLD learns retargeting with deep RL, with command-following delay below 0.11\,s~\cite{dep_efold}; DexTwist recognizes tripod pinches and estimates the intended screw axis for twisting tasks~\cite{dep_dextwist}; and offline pipelines add temporal consistency and contact maps for data generation~\cite{dep_dexflow}. Hardware-agnostic frameworks aim at ``any arm, any hand, any input device''~\cite{dep_openteledex,anyteleop}.

\paragraph{Haptic feedback.} Feedback improves demonstrations, but the evidence base is small. Early gripper studies found that haptic feedback raised door-opening collection throughput by 6\% and the success of resulting policies by about 11\%~\cite{dep_cuanhaptic}. On a dexterous hand, ViHaTeleop's US\$550 wrist-and-camera tracker with finger-wise vibrotactile feedback raised success by 2.2--15.6 percentage points across six contact-critical tasks with nine participants~\cite{dep_vihateleop}. Richer feedback seems to matter: a capstan-driven glove rendering planar (two-axis) fingertip force reduced contact-force error compared with visual-only and single-axis feedback~\cite{dep_n2d}, and a 32-DoF tactile fingertip display reduced the deviation between teleoperated and natural trajectories by 29--79\%, compressing the state distribution of demonstrations in ways the authors link to better policy training~\cite{dep_missingtouch}. Bilateral frameworks combine retargeting, multi-scale haptics, shared control and compliant arms~\cite{dep_bilateral}, and datasets collected with haptic teleoperation and fingertip touch are appearing~\cite{dep_haptile}. No large-$N$ study yet links multi-finger haptic feedback to downstream policy performance.

\paragraph{Shared autonomy and interventions.} Controlling 20+ DoF taxes operators, so control is increasingly shared. One system lets the human guide the arm in VR while an autonomous grasping VLA drives the fingers with tactile feedback, reaching 90\% success including unseen objects~\cite{dep_sharedvla}; joint human--agent learning adjusts the control ratio while data are collected~\cite{dep_hajl}. For corrections, abrupt takeovers cause ``gesture jumps'' in high-DoF hands; Hand-in-the-Loop blends human intent with policy execution, reducing intervention jitter by 99.8\% and grasp failures by 87.5\%, and policies trained on its corrections outperform those trained on standard teleoperation data by 19\%~\cite{dep_handitl}. These interfaces are the dexterous counterpart of DAgger-style intervention systems~\cite{dittox,pol_robocopilot}.

\subsection{Wearable, robot-free capture}
\label{sec:data-wearable}

In robot-free capture the human performs the task with their own hands while wearing hardware that makes the recording look like robot data, extending the handheld-gripper idea of UMI~\cite{umi} to multi-fingered hands. This was the fastest-growing route in 2025--2026.

\paragraph{Gloves plus wrist cameras.} DexCap pioneered portable motion capture with a chest-mounted RGB-D and SLAM camera~\cite{dexcap}. DexWild combines a glove, palm-mounted cameras mirrored on the robot, and single-camera wrist tracking; collectors gathered 9{,}290 demonstrations in 93 environments at 4.6$\times$ teleoperation speed, and co-training at a robot-to-human ratio of 1:2 reached 68.5\% success in unseen environments, about four times robot-only training~\cite{dexwild}. DexViTac adds high-resolution touch and collects more than 248 demonstrations per hour~\cite{dexvitac}.

\paragraph{Hand-specific exoskeletons.} DexUMI constrains the human hand to the target hand's workspace with a custom exoskeleton and removes the human hand from images by inpainting~\cite{dexumi}. DEXOP's ``perioperation'' paradigm mechanically links a passive hand exoskeleton to a passive robot hand, so the human directly drives the robot hand with natural force feedback; per task it was 2.4--8$\times$ faster than teleoperation~\cite{dexop}. Wearable-first designs such as WHED~\cite{whed} and DexEXO, which fits hand lengths of 140--217\,mm without calibration and puts a passive demonstration hand in the wrist camera's view so that no inpainting is needed~\cite{dexexo}, focus on comfort and cross-operator use.

\paragraph{The deployed hand as the collector.} The most direct way to eliminate the gap is to collect with the very hand that will be deployed. RealDexUMI wears an 11-DoF (6 active) hand with in-hand camera and fingertip touch on the human hand, with a palm-side isomorphic glove; it reports 88.75\% average success on eight tasks, falling to 70\% without touch, and one checkpoint ran on three different arms~\cite{realdexumi}. DexPort~\cite{dexport}, DEX-PHD~\cite{dexphd}, DEX-Mouse~\cite{dexmouse} and the co-designed Koala gripper~\cite{koala} follow similar ideas. The trade-off is that the data are bound to one hand design.

\paragraph{Scaled wearable programs.} Several programs operate at scale. XRZero-G0 replaces visual SLAM with VR tracking and closes a collect--check--train--evaluate loop with an 85\% valid-data rate over more than 2{,}000 hours of collection~\cite{xrzero}. World In Your Hands (WIYH) releases more than 1{,}000 hours with tactile signals and automatic annotation~\cite{wiyh}. YUBI's finger-aligned two-finger interface yields 8{,}434 hours and 1.2M episodes~\cite{yubi}. AdvDex/OmniShare uses a 29-encoder glove with Hall tactile arrays to collect more than 100k trajectories~\cite{advdex}. In industry, Sunday converts data from US\$200 gloves whose geometry and sensors match its robot hand into robot data, and reports about ten million episodes from more than 2{,}000 contributors~\cite{sundayact1,sunday36kr}; Generalist collects with wearable pincer-style ``data hands'', with press reports of growth from about 270k to more than 500k hours~\cite{gen15,generalistprofile}; Liber-0 uses a 131\,g, 0.3\,mm skin-colored glove with egocentric and two wrist cameras~\cite{liber0}; and Genesis AI pairs a sensorized glove with a hand that mirrors it~\cite{genesisai}. Collection-hardware product families are also appearing on the market~\cite{fastumi}.

\subsection{Egocentric human video}
\label{sec:data-ego}

Egocentric data record humans performing everyday tasks with head-mounted cameras, glasses or headsets; actions are recovered from estimated 3D hand poses. General-purpose corpora such as Ego4D~\cite{ego4d}, Ego-Exo4D~\cite{egoexo4d} and HOT3D~\cite{hot3d} provide limited hand annotation. EgoDex (829\,h, 194 tasks) records native 3D hand and finger tracking from Apple Vision Pro and is the most hand-relevant public dataset~\cite{egodex}. Headset-based human data have also been co-trained directly with humanoid data~\cite{humanpolicy}, and Figure pretrains on egocentric video collected in real homes~\cite{figuregobig}. Multi-institution platforms such as EgoVerse aggregate more than 1{,}300 hours~\cite{egoverse}. At the pretraining scale, EgoScale uses 20{,}854 hours of action-annotated egocentric video~\cite{egoscale}, UniHand-2.0 more than 35{,}000 hours across humans and 30 robots~\cite{beingh05}, and DreamDojo about 44k hours for world modeling~\cite{dreamdojo}. Meta's Aria Gen~2 glasses provide on-device hand, head and eye tracking for researchers~\cite{ariagen2}. Egocentric data also serve as language-grounded pretraining for the high-level ``brain'': PhysBrain converts egocentric video into large-scale VQA supervision~\cite{physbrain}.

Converting egocentric video into robot data involves five steps: (1)~hand reconstruction (native headset tracking, gloves, or monocular MANO estimation~\cite{mano,hamer} combined with camera SLAM); (2)~retargeting the wrist pose and fingertips with joint-limit and collision constraints~\cite{geort,anydexrt,toporetarget}; (3)~visual alignment by masking or inpainting the human hand and rendering the robot~\cite{egomimic,unidex}; (4)~physical correction in simulation~\cite{egoengine,spider}; and (5)~co-training with paired and robot data. Geometric plausibility does not imply dynamic feasibility: in EgoEngine's ablation, training directly on human video gave 0.03 success, adding an action branch 0.43, and the full pipeline 0.51~\cite{egoengine}. 2026 data engines therefore emphasize contact-aware and object-centric conversion from a single egocentric clip, followed by simulated RL and real deployment: EgoAERO checks the physical usability of a reconstruction online and decides at capture time whether to keep, repair or re-record~\cite{egoaero}; EgoMatrix~\cite{egomatrix}, EgoCPT~\cite{egocpt} and HandMimicX~\cite{handmimicx} treat human motion as a spatial prior and search for robot-native contacts; DexRoam captures whole-body egocentric data without external tracking~\cite{dexroam}; and Human2Dex separates human semantics from embodiment-specific execution~\cite{human2dex}.

\begin{figure*}[t]
\centering
\includegraphics[width=\textwidth]{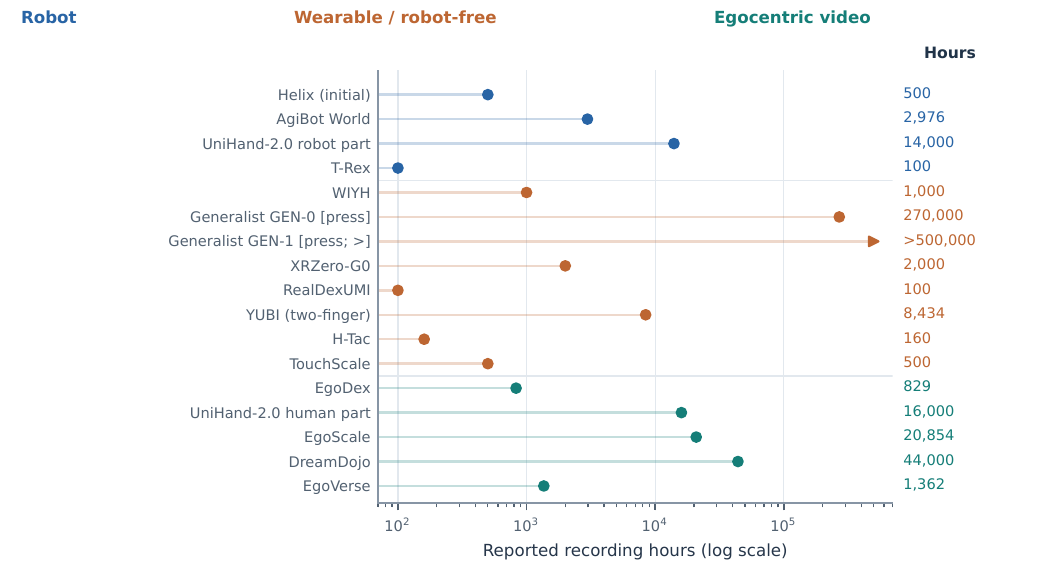}
\caption{Source-grouped reported recording hours, transcribed from the reviewed sources~\cite{figurehelix02,agibotworld,beingh05,trex,wiyh,generalistprofile,xrzero,realdexumi,yubi,htac,touchscale,egodex,egoscale,dreamdojo,egoverse}. Values are rounded where the source is approximate; the Generalist GEN-1 marker is a lower bound. Recording hours do not measure annotation quality, contact coverage or effective training exposure. Corpora can overlap and must not be summed. AgiBot World includes grippers; YUBI and Generalist use two-finger/pincer interfaces. This is a scale inventory, not a temporal scaling law.}
\label{fig:datascale}
\end{figure*}

\subsection{Simulation, generation and automation}
\label{sec:data-sim}

Simulation multiplies scarce real demonstrations: DexMimicGen generates large numbers of bimanual dexterous trajectories from a few human demonstrations~\cite{dexmimicgen}. Human-to-simulation-to-robot pipelines make human demonstrations physically feasible~\cite{maniptrans,human2sim2robot,dexmachina,spider}, and video generation and editing replace the human hand with a robot hand in pixels~\cite{egoengine,robotryon}. Fully automatic real-world collection is emerging for structured tasks: AutoDex runs unattended grasp collection with perception, execution, annotation and reset, at 4.8$\times$ teleoperation throughput~\cite{autodex}. Kinesthetic teaching with high-dimensional touch is another source of contact-rich labels~\cite{dextac}. We view these sources as amplifiers of real contact data rather than replacements.

\subsection{What the scaling evidence says}
\label{sec:data-scaling}

Several 2025--2026 results allow tentative conclusions about how data sources should be combined.

\begin{enumerate}
\item \textbf{Human data scale log-linearly, but need a bridge.} EgoScale reports a log-linear relation between hours of egocentric human data and validation loss, and task completion rising from 0.30 to 0.71 as human pretraining grows from 1k to 20k hours; a two-stage recipe (human pretraining, then mid-training on about 50 hours of aligned human data and 4 hours of robot data) improved average success by 54\% over no pretraining and transferred to a three-finger hand~\cite{egoscale}.
\item \textbf{Measured hand motion beats raw video volume.} Under identical post-training, Liber-0 reports that 10k hours of glove data outperform 2M hours of web human video on both pick-and-place and non-pick-and-place tasks; glove data plus 25\% of the robot demonstrations surpassed 100\% robot demonstrations (75.6 vs.\ 72.8 progress score); and glove data taught skills absent from robot data (non-pick-and-place progress rising from 2.2 to 47.5)~\cite{liber0}. These are blog-reported results without public code.
\item \textbf{Diversity of collectors and environments matters.} With a fixed number of glove demonstrations, increasing the number of collectors from 1 to 100 raised glove-only performance from 12.7 to 40.1~\cite{liber0}; DexWild finds similar benefits from environment diversity~\cite{dexwild}.
\item \textbf{Robot data remain the anchor.} DexWild finds that a 1:5 robot-to-human ratio is worse than 1:2~\cite{dexwild}; UniDex reports near-zero success without any robot data and that about two human demonstrations are worth one robot demonstration~\cite{unidex}. Physical Intelligence finds that human-to-robot transfer \emph{emerges} as the robot-data diversity of VLA pretraining grows~\cite{pih2r}.
\item \textbf{Human-likeness of the robot amplifies human data.} Liber-0's robot, with a three-DoF human-like wrist and human-aligned chest and wrist cameras, outperformed a conventional arm by 28.3 points on average given the same glove data~\cite{liber0}.
\end{enumerate}

\begin{figure*}[t]
\centering
\includegraphics[width=\textwidth]{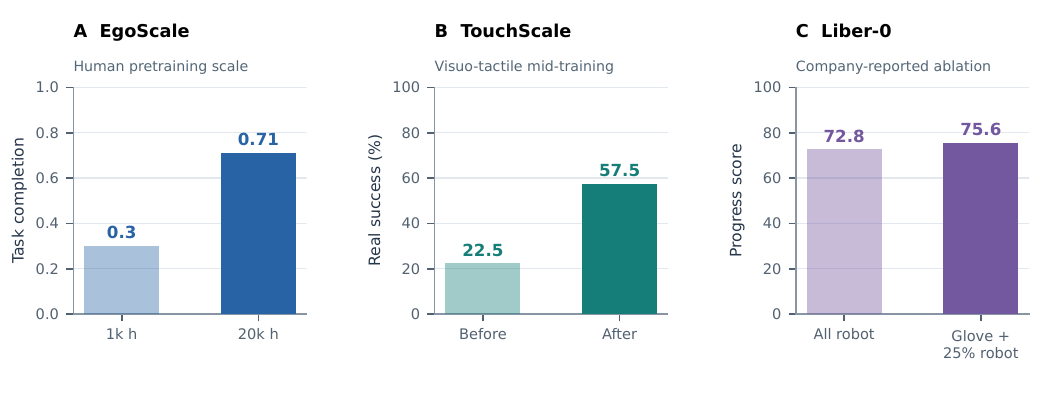}
\caption{Three within-study comparisons with deliberately separate metrics and axes. (A) EgoScale reports task completion at two human-pretraining scales~\cite{egoscale}; (B) TouchScale reports real success before/after visuo-tactile mid-training on four contact tasks~\cite{touchscale}; (C) Liber-0 reports progress with all robot demonstrations versus glove data plus 25\% of robot demonstrations~\cite{liber0}. Panel C is company-reported. Aggregate values are reproduced from the reviewed reports; underlying trial-level data are not provided here, so no uncertainty estimates or statistical significance are implied.}
\label{fig:evidence-panels}
\end{figure*}

\paragraph{What the comparisons do not establish.} \Cref{fig:evidence-panels} makes the metric distinction explicit. More human hours, a tactile mid-training stage and a changed data mixture are different interventions. Their effects cannot be pooled, and none identifies a universally optimal source mix across hands. A useful follow-up would vary one factor while holding target embodiment, total training compute, collector diversity and test objects fixed.

\subsection{Tactile data}
\label{sec:data-tactile}

Tactile data were long described as the largest gap, and 2026 saw rapid growth. TouchScale releases 500 hours (about 87k episodes) of bimanual full-palm tactile glove data synchronized with egocentric and wrist video; it reports that visuo-tactile mid-training raised real success on four contact tasks from 22.5\% to 57.5\% and that zero-shot contact prediction on an unseen tactile sensor improved from 0.134 to 0.383 IoU~\cite{touchscale}. Human-centric tactile pretraining corpora reach 160 hours~\cite{htac}; T-Rex collects 100 hours of tactile teleoperation~\cite{trex}; TactiDex aligns full-hand touch with kinematics and object state for 700+ sequences~\cite{tactidex}; and ART-Glove explores articulated tactile gloves that preserve human dexterity~\cite{artglove}. A 0.38 IoU on an unseen sensor is far from direct reuse; using the same tactile sensor at collection and deployment~\cite{dexumi,realdexumi} remains the safest choice.

\subsection{Processing toolchain}
\label{sec:data-tools}

Turning raw recordings into training data requires a chain of tools whose errors accumulate. \emph{Hand pose} comes from native headset tracking (Vision Pro in EgoDex~\cite{egodex}; on-device hand, head and eye tracking on research glasses~\cite{ariagen2}), from gloves (electromagnetic gloves output a 25-node skeleton and fingertip positions without optical occlusion~\cite{manusgloves}), or from monocular estimation of MANO parameters~\cite{mano,hamer}, whose translation drifts and is usually corrected by dynamic-scene reconstruction and camera SLAM. \emph{Retargeting} ranges from vector-optimization methods in the DexPilot tradition~\cite{dexpilot,anyteleop} to learned and calibration-free retargeters~\cite{geort,anydexrt}, fingertip IK with a virtual floating base and iterative correction for coupled joints~\cite{unidex}, and topology-preserving methods that handle penetration for downstream RL~\cite{toporetarget}. \emph{Physical repair} reconstructs the scene, corrects contacts with optimization or RL in simulation and renders robot observations~\cite{egoengine,egomatrix,egocpt,egoaero,spider}. The trend is from ``copying the human hand'' to ``using the human hand as a prior and searching for robot-native contacts''. \emph{Synchronization} of cameras, gloves, touch and wrist trackers must be at the millisecond level; AdvDex reports microsecond-level synchronization of a 29-encoder glove, Hall-effect touch and multi-view cameras~\cite{advdex}. Electromagnetic gloves lose accuracy near motors and metal and need on-site calibration.

\begin{table}[t]
\centering\tabsize
\caption{Recommended raw-data record for dexterous collection, so that retargeting and representations can be recomputed later (our recommendation, synthesized from the reviewed pipelines).}
\label{tab:datarecord}
\begin{tabularx}{\columnwidth}{L{2.0cm}Y}
\toprule
Field & Content and rationale \\
\midrule
Hand state & MANO 21 keypoints and/or glove joint angles at native rate; keep raw, not retargeted~\cite{mano,egoscale} \\
Wrist pose & 6D pose with tracker type and calibration; main source of action noise~\cite{dexumi,dexop} \\
Vision & Raw egocentric, wrist and palm streams with intrinsics/extrinsics; mirrored camera placement on robot~\cite{dexwild,liber0} \\
Touch & Raw taxel signals with sensor model, layout and calibration; optionally a MANO-UV projection~\cite{unitachand} \\
Objects & Meshes and 6D poses where available; required by physics-aware retargeting~\cite{spider,omnihoi,contactanchored} \\
Robot data & Commands, measured joint states, currents/torques, control rate and latency~\cite{beingh05} \\
Annotations & Sub-task segmentation, language, success/failure, takeover reason, quality score~\cite{curationmetrics} \\
Provenance & Collector ID (pseudonymized), site, consent record, de-identification status~\cite{miitstandard} \\
\bottomrule
\end{tabularx}
\end{table}

\subsection{Collection hardware and the data industry}
\label{sec:data-industry}

\paragraph{Device costs.} Collection hardware spans three orders of magnitude. Research electromagnetic gloves cost roughly US\$5k--6.3k per pair, and around US\$7.5k--9.7k with haptic feedback, with lead times of weeks~\cite{manusgloves}; a force-feedback glove with palm haptics launched at US\$5{,}999~\cite{senseglovenova}; research glasses with on-device tracking are distributed by application~\cite{ariagen2}; at the low end, hardware-matched gloves sell for about US\$200~\cite{sunday36kr}, and open designs have BOMs below US\$500 for a tactile-feedback glove~\cite{feelrobot} and below US\$150 for a handheld interface~\cite{dexmouse}.

\paragraph{Data factories and markets.} In China, data collection is becoming an industry. Large teleoperation data centers produce standardized multi-embodiment datasets~\cite{agibotworld}; tactile-sensor companies operate clusters of collection ``factories'' and a data cloud marketplace that advertises high-DoF human-hand data with six-axis touch~\cite{paxinimall,paxinipxcap}; robot-free collection product families are sold through retail channels and paired with million-hour collection plans~\cite{fastumi}; companies pursuing human data report exoskeleton tactile gloves with in-the-wild pipelines and targets of a million hours~\cite{psibotmanual,psir2}; and the first embodied-intelligence dataset, containing video, joint angles and torques, was listed and traded on a provincial data exchange in January 2026~\cite{datatrade}. Estimates of the supply--demand gap vary widely: one provincial data bureau put compliant real-interaction data in China at about 500k hours against a commercial need of tens of millions~\cite{datashortage}. Public research institutes advocate classification, grading and de-identification at the source and authorization, notarization and traceability at sharing time for industrial manipulation data~\cite{caictdata}.

\subsection{Quality, cost and governance}
\label{sec:data-governance}

\paragraph{Curation.} As volumes grow, deciding what to keep sets the policy ceiling. Demo-SCORE trains a quality classifier from policy rollouts~\cite{demoscore}; DemInf scores demonstrations by state--action mutual information~\cite{deminf}; SCIZOR removes suboptimal and redundant state--action pairs~\cite{scizor}; influence functions estimate each trajectory's contribution~\cite{qoq}; and smoothness metrics provide zero-training filters~\cite{smoothcuration}. A systematic evaluation cautions that metrics that detect bad demonstrations do not necessarily yield better policies, so curation should be judged by downstream policy performance~\cite{curationmetrics}. In practice, \emph{online} quality checks at capture time~\cite{egoaero,xrzero} keep bad data out of the corpus.

\paragraph{Cost.} Industry estimates for China put real-robot teleoperation at roughly CNY~500--1{,}000 per hour, egocentric capture at CNY~100--200 per hour, and UMI-style capture at US\$0.6--1.2 per episode~\cite{costcsdn}; these are order-of-magnitude figures from secondary sources. A crowdsourced egocentric-video gig economy has emerged, with pay from about US\$1 to US\$15 per hour depending on region and payment typically per \emph{accepted} hour~\cite{techreviewgig,humanarchive}.

\paragraph{Standards and ethics.} China's first quality standard for embodied-intelligence datasets, with eight quality dimensions (completeness, consistency, diversity, authenticity, usability, utility, extensibility and security), took effect on 1 November 2026~\cite{miitstandard}, and an ISO work item on humanoid robot datasets was approved in 2026~\cite{isostandard}. Home egocentric video exposes interiors and habits even without faces; reports document collectors who are unsure how their data are stored or shared~\cite{techreviewgig,humanarchive}. Informed consent, de-identification and deletion procedures should be designed in from the start, both for ethical reasons and to keep data tradable and reusable.

\takeaway{Several systems support a pyramid recipe: large-scale egocentric pretraining, tens of hours of paired human--robot mid-training, and a small amount of target-robot teleoperation for post-training. What decides success is less the capture device than kinematic, visual and tactile alignment between human and robot, and the quality-control pipeline.}

\section{Human Hands as a Data Source}
\label{sec:human}

Because human data now dominate dexterous pretraining (\cref{sec:data}), the computer-vision infrastructure for perceiving human hands has become part of the robot-learning stack. This section reviews hand and hand--object perception, the human interaction datasets that underpin conversion pipelines, and the lineage of methods that learn robot dexterity from human video.

\subsection{Perceiving human hands}
\label{sec:human-perception}

\paragraph{3D hand pose.} Parametric hand models, chiefly MANO~\cite{mano}, provide the common output space. Early monocular systems such as FrankMocap regressed hand and body motion jointly at about 9.5\,fps~\cite{hum_frankmocap}; transformer-based reconstruction (HaMeR~\cite{hamer}) and WiLoR, which couples a real-time fully convolutional hand localizer with transformer reconstruction trained on more than 2M in-the-wild images and yields smooth tracking without temporal modules~\cite{hum_wilor}, are today's default front ends.

\paragraph{World-frame and 4D tracking.} Egocentric videos involve a moving camera, so per-frame camera-space poses must be lifted to a world frame to serve as wrist actions. HaWoR separates camera-space hand motion from the camera trajectory using adaptive egocentric SLAM and infills frames where hands leave the view~\cite{hum_hawor}; Dyn-HaMR recovers 4D global motion of interacting hands from dynamic-camera video by multi-stage optimization with an interacting-hand prior~\cite{hum_dynhamr}. Low-cost ground truth is obtained with back-mounted multi-camera rigs combined with headset cameras~\cite{hum_egoexo3d}. World-frame tracking accuracy is, in our assessment, the main perception bottleneck for turning egocentric video into actions.

\paragraph{Hand--object reconstruction and contact.} Robot conversion needs the object as well as the hand. HOLD reconstructs articulated hands and unknown objects from monocular interaction video without 3D annotations~\cite{hum_hold}; MagicHOI uses novel-view diffusion priors to supervise unseen object regions in short clips~\cite{hum_magichoi}; and G-HOP learns a category-conditioned diffusion prior over hand--object interaction from seven datasets spanning 155 categories, usable for both reconstruction and grasp synthesis~\cite{hum_ghop}. Contact supervision comes from thermal imaging (ContactDB, 3{,}750 contact maps over 50 objects~\cite{hum_contactdb}; ContactPose, 2{,}306 grasps from 50 participants~\cite{hum_contactpose}) and from dense vertex-level contact estimators that address the strong imbalance between contact and non-contact regions~\cite{hum_haco}. These contact signals are exactly what contact-anchored retargeting needs (\cref{sec:ce-data}), yet they are still rarely used in robot pipelines.

\subsection{Hand--object interaction datasets}
\label{sec:human-datasets}

\Cref{tab:hoidata} summarizes human hand--object datasets used for retargeting and pretraining. Classic datasets are small in object count and duration but precise: FPHA introduced first-person RGB-D with magnetic-sensor hand poses~\cite{hum_fpha}; HO-3D annotated 77{,}558 frames markerlessly~\cite{hum_ho3d}; DexYCB benchmarks grasping of YCB objects and human-to-robot handover~\cite{hum_dexycb}; GRAB captures whole-body grasping with contact~\cite{hum_grab}. Category-level and bimanual datasets followed: H2O~\cite{hum_h2o}, HOI4D with 2.4M egocentric frames over 800 object instances~\cite{hum_hoi4d}, OakInk with affordance-aware interactions over 1{,}800 objects~\cite{hum_oakink}, OakInk2~\cite{oakink2}, ARCTIC~\cite{arctic}, TACO for bimanual tool use~\cite{hum_taco}, Assembly101 with 18M 3D hand poses over 4{,}321 procedural videos~\cite{hum_assembly101}, HOT3D~\cite{hot3d}, GigaHands~\cite{gigahands} and HOGraspNet, which labels the full grasp taxonomy and contact maps over 1.5M frames from 99 participants~\cite{hum_hograspnet}. Recent work moves to in-the-wild scenes, including outdoors~\cite{hum_show3d}. Several of these are the raw material of robot-data conversion: UniDex builds its 8-hand corpus from H2O, HOI4D, HOT3D and TACO~\cite{unidex}, and DexMachina and contact-anchored retargeting start from captured human--object interaction~\cite{dexmachina,contactanchored}. Compared with the tens of thousands of hours of robot-oriented egocentric data (\cref{fig:datascale}), however, precisely annotated HOI data remain orders of magnitude smaller.

\begin{table}[t]
\centering\tabsize
\caption{Human hand--object interaction datasets relevant to robot transfer (as reported).}
\label{tab:hoidata}
\begin{tabularx}{\columnwidth}{L{1.95cm}L{0.55cm}Y}
\toprule
Dataset & Year & Content / scale \\
\midrule
FPHA~\cite{hum_fpha} & 2017 & 100k+ RGB-D frames, 45 actions, magnetic hand pose \\
HO-3D~\cite{hum_ho3d} & 2019 & 77{,}558 frames, 10 objects, markerless 3D \\
ContactPose~\cite{hum_contactpose} & 2020 & 2{,}306 grasps, thermal contact maps \\
GRAB~\cite{hum_grab} & 2020 & Whole-body mocap grasps, 51 objects, contact \\
DexYCB~\cite{hum_dexycb} & 2021 & Multi-view YCB grasps; handover benchmark \\
HOI4D~\cite{hum_hoi4d} & 2022 & 2.4M ego frames, 800 instances, 16 categories \\
OakInk~\cite{hum_oakink} & 2022 & 1{,}800 objects w/ affordances; 50k interactions \\
Assembly101~\cite{hum_assembly101} & 2022 & 4{,}321 videos, 18M 3D hand poses \\
TACO~\cite{hum_taco} & 2024 & 2.5k bimanual tool-use sequences \\
HOGraspNet~\cite{hum_hograspnet} & 2024 & 1.5M frames, full grasp taxonomy, contact \\
SHOW3D~\cite{hum_show3d} & 2026 & In-the-wild scene-level 3D HOI \\
\bottomrule
\end{tabularx}
\end{table}

\subsection{Learning robot dexterity from human video}
\label{sec:human-video}

The idea of learning dexterity from human video predates the foundation-model era, and its evolution explains today's designs. DexMV recorded human videos of the same tasks as a multi-finger simulation, extracted 3D hand--object poses and converted them into robot demonstrations that solved tasks RL alone could not~\cite{hum_dexmv}; VideoDex used visual, action and physical priors from internet hand-use video to guide an arm--hand system~\cite{hum_videodex}; and MimicPlay learned latent plans from human play data to condition low-level robot policies~\cite{hum_mimicplay}. Two families have since emerged.

\paragraph{Lift to 3D, then retarget or reinforce.} ViViDex uses reconstructed human trajectories as reward guidance for RL and distills state policies into point-cloud policies~\cite{hum_vividex}; HOP pretrains a hand--object interaction policy from video for later fine-tuning~\cite{hum_hop}; DexH2R learns a residual policy on top of retargeted human motion and reports a 40\% gain over prior state of the art across settings, without teleoperation~\cite{hum_dexh2r}; VIDEOMANIP reconstructs hand--object trajectories from monocular RGB without capture devices, reaching 62.86\% average real success over seven LEAP Hand tasks, outperforming retargeting-based baselines by 15.87\%~\cite{hum_videomanip}; and X-Sim builds a photorealistic simulation from RGB-D human video, trains RL with object-centric rewards and matches behavior cloning with 10$\times$ less collection time~\cite{hum_xsim}. Embodiment-agnostic intermediate representations avoid joint-level correspondence altogether: Motion Tracks predicts 2D tracks lifted to 6D as a shared human--robot action space, reaching 86.5\% average success over four real tasks with a gripper-type end effector~\cite{hum_motiontracks}; Dexterous Point Policy predicts 3D object and hand keypoints autoregressively from human video and reports 75.0\% real success where a VLA baseline achieves 1.0\%~\cite{hum_dexpointpolicy}. The Motion Tracks result uses a gripper-type end effector, so it is evidence for an intermediate representation rather than a direct multi-fingered-hand result.

\paragraph{Edit pixels.} Phantom estimates hand pose, inpaints the human arm and overlays a rendered robot, enabling zero-shot policies without robot data with up to 92\% success on deformable-object, sweeping and insertion tasks~\cite{hum_phantom}; Masquerade edits 675k egocentric frames into bimanual robot views to pretrain visual encoders, gaining 5--6$\times$ over baselines with log-scaling in edited data~\cite{hum_masquerade}. These are precursors of the generative ``robot try-on'' methods discussed in \cref{sec:ce-data}~\cite{egoengine,robotryon}.

\paragraph{Glasses and VLAs.} Research glasses with on-device hand tracking simplify capture: EgoZero trains gripper policies (on a Franka arm) from about 20 minutes of glasses data per task with 70\% zero-shot success~\cite{hum_egozero}, and AINA extends this to multi-fingered hands with no robot data, simulation or RL across nine everyday tasks~\cite{hum_aina}. EgoVLA predicts wrist and hand actions from egocentric video and fine-tunes on a few robot demonstrations~\cite{hum_egovla}; web-video skill policies for gripper robots cover open, close, pour, cut and stir families zero-shot~\cite{hum_zeromimic}; and co-training human video with robot data nearly doubles performance on generalization settings present only in human data, but only once pretraining is sufficiently diverse~\cite{hum_pihuman,pih2r}.

\paragraph{Hand motion priors.} Generative models of hand--object interaction provide priors and augmentation: text-to-HOI generation via contact maps~\cite{hum_text2hoi}, two-stage grasp-then-manipulate diffusion~\cite{hum_diffh2o}, hand synthesis from object trajectories~\cite{hum_manidext}, and FlowHOI, which generates hand poses, object poses and contact states from egocentric observations and language with flow matching and executes them on real dexterous robots~\cite{hum_flowhoi}. Closing the loop from such generators to robot execution remains rare.

\takeaway{Human-hand perception has become robot infrastructure. The weakest links are world-frame hand tracking in egocentric video, object reconstruction from short clips and the scarcity of contact labels. Methods split between lifting to 3D (then retargeting, RL or keypoint policies) and editing pixels; the former gives control over physics, the latter over appearance, and the strongest recent systems combine both.}

\section{Layer 3: Cross-Embodiment Transfer}
\label{sec:crossemb}

Two gaps overlap in dexterous manipulation. The \emph{human-to-robot} gap separates a human hand with about 27 DoF, compliant skin and full-hand touch from robot hands with 6--24 active DoF, rigid links and different proportions; human video does not directly provide executable actions, the hand looks different in images, and contact points do not correspond. The \emph{robot-to-robot} gap arises because hands differ in finger count (three to five), DoF, coupling, joint ordering, thumb design and actuation, so action vectors do not share dimension or meaning. Both are compounded by differences in camera and tactile configuration, dynamics, and timing. Being-H0.5, for instance, introduces a dedicated asynchronous chunking scheme because embodiments with different latencies and control rates otherwise produce jittery chunk transitions~\cite{beingh05}. Solutions operate at three levels---data, representation and model (\cref{tab:crossemb})---and practical systems combine them.

\begin{table*}[t]
\centering\tabsize
\caption{Cross-embodiment approaches for dexterous hands. H$\to$R: human-to-robot; R$\leftrightarrow$R: robot-to-robot. Numbers are as reported under each paper's own protocol.}
\label{tab:crossemb}
\begin{tabularx}{\textwidth}{L{2.6cm}L{1.1cm}L{4.4cm}L{2.0cm}Y}
\toprule
Approach & Gap & Representative work & Contact/force & Reported evidence \\
\midrule
Kinematic retargeting + video & H$\to$R & EgoScale~\cite{egoscale}, UniDex~\cite{unidex}, Do as I Do~\cite{doasido} & No & +54\% from 20k\,h pretraining; 8 hands, 52k trajectories \\
Physics/contact-aware retargeting & H$\to$R & SPIDER~\cite{spider}, OmniHOI~\cite{omnihoi}, contact-anchored~\cite{contactanchored}, TopoRetarget~\cite{toporetarget} & In sim & 9 embodiments; 5 hands 39--89\% vs.\ $\le$31\%; 7.3$\to$59.3\% \\
Generative visual conversion & H$\to$R & EgoEngine~\cite{egoengine}, RoboTryOn~\cite{robotryon}, DexStitch~\cite{dexstitch} & No & Zero real-demo dexterous policy (EgoEngine) \\
Deployed-hand / isomorphic capture & H$\to$R, arms & DexUMI~\cite{dexumi}, RealDexUMI~\cite{realdexumi} & Yes & 88.75\%; one checkpoint on 3 arms \\
Sim RL grounding & H$\to$R & ManipTrans~\cite{maniptrans}, Human2Sim2Robot~\cite{human2sim2robot} & In sim & One human demo to real \\
Unified slot action space & R$\leftrightarrow$R & FAAS~\cite{unidex}, DexFormer~\cite{dexformer}, Being-H0.5~\cite{beingh05}, AcrossDEX~\cite{acrossdex} & No & 30 robots, one checkpoint on 5 platforms \\
Relative EEF + adapters & both & GR00T N1.7~\cite{gr00tn17}, Being-H0~\cite{beingh0} & No & Contested by $\pi_{0.7}$~\cite{pi07} \\
Canonical / aligned morphology & R$\leftrightarrow$R & One Hand~\cite{onehand}, DexGrasp-Zero~\cite{dexgraspzero} & No & 82\% real zero-shot grasping \\
Contact / tactile representations & both & MARC~\cite{transfercontact}, UniTacHand~\cite{unitachand}, TactAlign~\cite{tactalign}, GOTT~\cite{gott} & Core & Tactile zero-shot H$\to$R \\
Latent action / world models & both & OPFA~\cite{opfa}, LDA-1B~\cite{lda1b}, DreamZero~\cite{dreamzero}, DexWM~\cite{dexwm} & Indirect & 8 demos $\approx$ 72; 30-min adaptation \\
\bottomrule
\end{tabularx}
\end{table*}

\begin{figure*}[t]
\centering
\includegraphics[width=\textwidth]{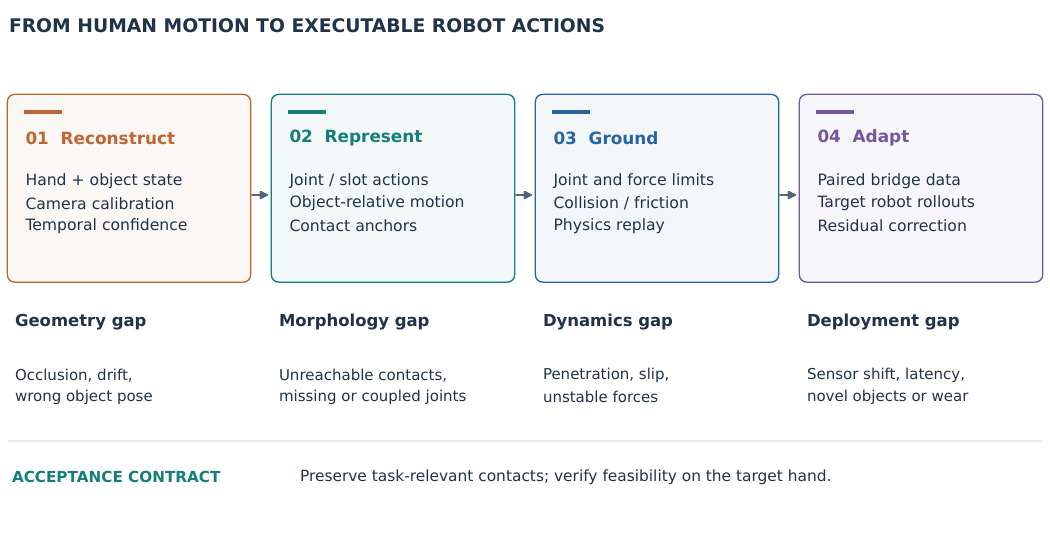}
\caption{From human observations to target-hand actions: a conceptual synthesis of reconstruction, representation, physics grounding and adaptation~\cite{egoscale,unidex,spider,contactanchored}. The lower row identifies the failure each stage must diagnose. Not every method implements every stage, and physics replay is a feasibility test under a model rather than proof of real-world safety.}
\label{fig:transfer-pipeline}
\end{figure*}

\Cref{fig:transfer-pipeline} separates four transformations that are often conflated under ``retargeting''. Preserving a wrist trajectory, reproducing an image and maintaining stable object contact are different objectives. A useful comparison must specify which objective is optimized and which constraints are tested on the target embodiment.

\subsection{Data level: making human data executable}
\label{sec:ce-data}

\paragraph{Kinematic retargeting.} Retargeting maps human keypoints or glove joint angles to robot joints and underlies almost every pipeline. The classic approach optimizes fingertip and keypoint vectors~\cite{dexpilot,anyteleop}; GeoRT learns an unsupervised neural retargeter running at 1\,kHz~\cite{geort}; AnyDexRT needs no calibration~\cite{anydexrt}; and CrossDex approximates retargeting with a network to speed up RL~\cite{crossdex}. EgoScale found joint-space retargeted actions to be a better pretraining representation than wrist-only or fingertip-only actions~\cite{egoscale}.

\paragraph{From ``looking similar'' to ``making the same contact''.} The 2026 trend is contact- and physics-preserving retargeting. TopoRetarget builds a hand--object interaction mesh and solves a Laplacian optimization, reducing contact error to 7.71\,mm and raising downstream RL pen-spinning success from 46.9\% to 87.5\%~\cite{toporetarget}. SPIDER uses sampling-based optimization in parallel physics simulation with curriculum-style virtual contact guidance to turn kinematic human demonstrations into dynamically feasible trajectories for nine embodiments and releases thousands of simulated demonstrations~\cite{spider}. OmniHOI reconstructs hand--object interaction from monocular video and transfers it to five hands of 6--22 DoF with 39--89\% success, versus at most 31\% for prior transfer methods~\cite{omnihoi}. Contact-anchored retargeting argues that what survives the human-to-robot mapping is the \emph{contact structure} (which finger segment touches which part of the object, in what order) rather than joint motion; refining the human hand in simulation, retargeting to hand-agnostic contact anchors and learning a residual policy raised single-hand success from 7.3\% to 59.3\% across 25{,}454 trajectories and transferred one human dataset to four hands~\cite{contactanchored}. ManipTrans~\cite{maniptrans} and Human2Sim2Robot~\cite{human2sim2robot} use residual learning or RL in simulation to the same end.

\paragraph{Converting video at scale.} UniDex converts H2O, HOI4D, HOT3D and TACO into about 52k trajectories for eight hands (6--24 DoF), removing the human hand and rendering the robot hand into point clouds~\cite{unidex}. Do as I Do~\cite{doasido}, DexCast~\cite{dexcast}, Dex-X~\cite{dexx} and DexStitch~\cite{dexstitch} reconstruct interactions from everyday video, the latter segmenting long videos into composable skills. Generative methods close the \emph{visual} gap: EgoEngine generates robot-embodied observation videos with feasible action trajectories and reports a zero-shot deployable dexterous visuomotor policy trained without real robot demonstrations~\cite{egoengine}; RoboTryOn frames the conversion as virtual try-on, translating keyframes to a target robot and labeling actions with an inverse dynamics model~\cite{robotryon}. Generated pixels can, however, depict grasps that the robot cannot physically form, so feasibility checks remain necessary.

\paragraph{Paired datasets.} HRDexDB records humans and multiple robot hands grasping the same 100 objects with 3D reconstructions and success labels~\cite{hrdexdb}, a rare resource for studying human-to-robot contact transfer. Paired robot-to-robot data for the same tasks remain scarce.

\subsection{Representation level: a common language for hands}
\label{sec:ce-repr}

\paragraph{Function-aligned slot action spaces.} Joints with the same function share a dimension, and missing joints are masked. UniDex's FAAS uses 82 dimensions (18 for wrists, 64 for joints, of which 21 base slots are shared across hands)~\cite{unidex}; DexFormer uses a 20-D canonical finger space aligned by anatomical role and infers morphology from history rather than an embodiment ID~\cite{dexformer}; Being-H0.5 maps 30 robots and the human MANO hand into semantically aligned slots, treating the human hand as one more embodiment~\cite{beingh05}; AcrossDEX shares the wrist representation and distinguishes fingers with embodiment fields and validity masks~\cite{acrossdex}. A sphere-based unified hand action space with cascaded IK reports zero-shot transfer to unseen hands for in-hand reorientation~\cite{uhas}.

\paragraph{Canonical morphology.} \emph{One Hand to Rule Them All} parses any URDF into a canonical hand (up to five fingers and 22 DoF) and learns a 16-D morphology latent from 65{,}536 synthetic hands; a policy conditioned on it grasps with an unseen three-finger hand at 81.9\% success in simulation~\cite{onehand}. DexGrasp-Zero aligns graph-network policies to shared mechanical semantics and, trained on four hands, reaches 85\% zero-shot grasping on two unseen hands in simulation and 82\% on three real platforms~\cite{dexgraspzero}; this is among the strongest real zero-shot results between multi-fingered hands, albeit limited to grasping. The D(R,O) representation encodes grasps as robot--object distance matrices~\cite{drograsp}.

\paragraph{Human hand as intermediary.} CrossDex outputs human eigengrasp weights that are then retargeted to each robot~\cite{crossdex}; DexWM uses MANO keypoint displacements as actions and even builds virtual keypoints for parallel grippers~\cite{dexwm}.

\paragraph{Relative end-effector actions.} The wrist's relative pose change is the most easily shared part of an action. GR00T N1.7 adopts a relative end-effector (EEF) action space shared between humans and robots and attributes much of its cross-embodiment performance to it~\cite{gr00tgithub}; Being-H0 reaches a similar conclusion~\cite{beingh0}. $\pi_{0.7}$, in contrast, reports in supplementary experiments no clear benefit of EEF control for cross-embodiment transfer and uses joint space~\cite{pi07}. The embodiment sets differ, so the results need not conflict, but the choice of action space remains open.

\paragraph{Contact, force and touch.} Motion alone does not transfer contact. ``Transferring Contact, Not Just Motion'' pairs a glove motion latent with calibrated joint torques to derive fingertip force descriptors, enabling compliant grasping across hands even when vision is masked~\cite{transfercontact}.
\label{sec:ce-tactile}
For touch, UniTacHand projects glove and robot tactile signals onto the UV map of the MANO surface, a morphology-consistent 2D canvas, and reports zero-shot tactile policy transfer from human to robot with 10 minutes of paired data~\cite{unitachand}. TactAlign aligns human glove and robot touch with rectified flow without paired data, achieving zero-shot light-bulb screwing; removing alignment drops average success to 28\%~\cite{tactalign}. AnyTouch learns sensor-agnostic visuo-tactile features~\cite{anytouch}. GOTT is object-centric: given robot-agnostic object trajectories, a single cross-embodiment contact-acquisition primitive establishes stable contact and a pose-conditioned controller tracks the object~\cite{gott}.

\paragraph{Latent actions.} Latent actions learned from consecutive frames act as pseudo-labels for unlabeled video, with embodiment-specific input and output layers re-initialized at post-training~\cite{lapa,dreamdojo}. OPFA learns geometry-aware action latents with a unified decoder across 11 end-effectors; joint training improves success by more than 50\%, and about eight demonstrations on a new embodiment approach the result of 72~\cite{opfa}. Gemini Robotics 1.5's Motion Transfer shares one VLA across ALOHA, bimanual Franka and a humanoid without explicit action alignment~\cite{geminirobotics15}.

\subsection{Model level: one model, many hands}
\label{sec:ce-model}

The dominant pattern is a shared VLM backbone with an action expert (flow matching or diffusion) and lightweight embodiment-specific state encoders and action heads, as in EgoScale's GR00T-style model~\cite{egoscale,gr00tn1}, UniDex-VLA (point-cloud encoder on a $\pi_0$ backbone; 81\% average progress on tool use versus 38\% for $\pi_0$)~\cite{unidex,pi0} and GR00T N1.7~\cite{gr00tn17}. Being-H0.5 separates shared motion primitives from embodiment experts with a mixture-of-flow head to reduce negative transfer between embodiments~\cite{beingh05}. World-action models absorb unlabeled data and reduce dependence on a unified action space: LDA-1B assigns data roles by quality (clean trajectories train the policy, noisy ones only dynamics, unlabeled human video only visual prediction) and was evaluated on hands of 10 and 22 DoF~\cite{lda1b}; DreamZero adapts to a new robot with 30 minutes of play data and gains more than 42\% on unseen tasks from 10--20 minutes of human or other-robot video~\cite{dreamzero}. Morphology-randomized RL in simulation trains one policy across hundreds of hands~\cite{dexformer,onehand}. Finally, hierarchical systems keep the cross-embodiment ``what and where'' in the large model and leave embodiment-specific high-frequency contact control to a dedicated layer (\cref{sec:learn-hier}).

\subsection{A decision guide}
\label{sec:ce-guide}

\Cref{tab:ceguide} condenses the evidence of this section into recommendations by situation. Its entries are our synthesis and should be read as starting points for ablation, not as settled answers.

\begin{table}[t]
\centering\tabsize
\caption{Choosing a cross-embodiment strategy (our synthesis of the reviewed evidence).}
\label{tab:ceguide}
\begin{tabularx}{\columnwidth}{L{2.35cm}Y}
\toprule
Situation & Suggested starting point \\
\midrule
One target hand, budget for hardware & Isomorphic or deployed-hand capture with shared touch~\cite{realdexumi,dexumi,dexport}; add teleoperation only for failure-prone steps \\
Several hands, mostly grasping & Morphology-aligned or canonical-hand policies~\cite{dexgraspzero,onehand}; cross-hand grasp synthesis~\cite{sim_gendexgrasp} \\
Several hands, general tasks & Function-aligned slot actions with masks and embodiment adapters~\cite{unidex,beingh05}; ablate relative-EEF vs.\ joint wrist actions~\cite{gr00tn17,pi07} \\
Large unlabeled video & Latent actions or world-action pretraining~\cite{opfa,dreamzero,pol_univla}, then paired mid-training~\cite{egoscale} \\
Human video, contact-rich tasks & Contact-anchored or physics-aware retargeting with object meshes~\cite{contactanchored,spider,omnihoi}; residual policies on the robot \\
Visual gap dominates & Mirror camera placement~\cite{dexwild,liber0}; then inpainting or generative conversion with feasibility checks~\cite{egoengine,robotryon} \\
Touch must transfer & Same sensor at capture and deployment; otherwise canonical surfaces or unpaired alignment~\cite{unitachand,tactalign} \\
New robot, little data & Start from an open base model and test the ``30 minutes / ten demonstrations'' regime~\cite{gr00tn17,dreamzero,opfa} \\
\bottomrule
\end{tabularx}
\end{table}

\takeaway{The mainstream recipe is ``human data as the main ingredient, a unified or alignable hand action space, large-model pretraining, and a small amount of target-hand data''. Adaptation cost dropped from hours to minutes or a handful of demonstrations in 2026, but most zero-shot evidence is between grippers and arms; zero-shot transfer between multi-fingered hands is convincing mainly for grasping. Contact and touch are the next transfer anchor.}

\section{Touch Across the Stack}
\label{sec:touch}

Touch appears in every layer of our taxonomy: as sensor hardware (L1), as a data modality (L2), as a transfer anchor (L3) and as a policy input (L4). Because it is widely regarded as the decisive missing ingredient for contact-rich dexterity, and because the 2024--2026 literature on it grew very quickly, we devote a cross-cutting section to it. A companion review organizes tactile robot learning by the functional role of touch~\cite{tac_survey}; here we focus on what matters for multi-fingered hands.

\subsection{Sensors for hands}
\label{sec:touch-sensors}

\paragraph{Vision-based sensors.} Camera-based sensors image the deformation of an elastomer and offer the highest spatial resolution. DIGIT made such sensors compact and cheap enough to mount on multi-fingered hands and was used for in-hand marble manipulation~\cite{tac_digit}; Digit~360 is a multimodal fingertip with about 8.3M taxels, 7\,$\mu$m spatial features, about 1\,mN normal and shear force resolution, and sensing of vibration up to 10\,kHz, heat and odor~\cite{tac_digit360}; 9DTact reconstructs 3D shape and regresses 6D force without markers, trained on about 100k image--force pairs~\cite{tac_9dtact}. Their drawbacks for hands are size, curvature constraints, bandwidth and the cost of covering more than the fingertips.

\paragraph{Magnetic and resistive skins.} Magnetic skins separate electronics from a replaceable passive elastomer. ReSkin introduced this design with learning that is robust to batch variation~\cite{tac_reskin}; AnySkin made it plug-and-play and reported the first uncalibrated sensor whose learned policies generalize across sensor instances~\cite{tac_anyskin}; eFlesh 3D-prints magnetic cut-cell microstructures in arbitrary convex shapes for under US\$5 in magnets, with 0.5\,mm localization error, 95\% slip detection on unseen objects and a 40\% gain of visuo-tactile over vision-only policies~\cite{tac_eflesh}. Piezoresistive skins dominate mass-produced humanoid hands for cost reasons~\cite{eskinroutes}, and Hall-effect arrays provide dense coverage at moderate cost~\cite{paxinimall}.

\paragraph{Full-hand coverage.} F-TAC Hand embeds high-resolution sensing (0.1\,mm) over 70\% of the hand surface; in 600 real trials its tactile-equipped system significantly outperformed non-tactile alternatives~\cite{tac_ftac}. Pretrained encoders now exist for skins distributed over fingertips, phalanges and palm (Sparsh-skin, on an Allegro Hand with magnetic sensors), improving over prior work by more than 41\%~\cite{tac_sparshskin}. Industry has moved in the same direction (\cref{sec:hw-tactile}), and investment in tactile sensor companies surged in 2025--2026~\cite{tactilefunding}.

\paragraph{How much resolution is needed?} The evidence is split. Dense \emph{binary} contact over palm, links and fingertips suffices for blind in-hand rotation with zero-shot transfer~\cite{tac_touchdex}, while other work pushes resolution to millions of taxels~\cite{tac_digit360} or 0.1\,mm skin~\cite{tac_ftac}. A plausible resolution is task dependence: coarse, wide coverage for grasp stability and reorientation, and high resolution for slip, texture and sub-millimeter insertion.

\subsection{Simulating touch}
\label{sec:touch-sim}

Four strategies compete. \emph{Rendered optical simulation} reproduces vision-based sensor images: TACTO renders at hundreds of frames per second and supported grasp-stability learning on one million simulated grasps~\cite{tac_tacto}; Taxim calibrates an example-based GelSight model from fewer than 100 examples~\cite{tac_taxim}; and TacSL renders visuotactile images and force fields in Isaac more than 200$\times$ faster than prior work, with an asymmetric actor--critic distillation scheme for sim-to-real~\cite{tac_tacsl}. \emph{Physics-based} simulators model the elastomer: DiffTactile is differentiable and handles elastic, elastoplastic and cable-like objects~\cite{tac_difftactile}, and Taccel parallelizes incremental-potential contact to run 18$\times$ faster than real time across thousands of environments~\cite{tac_taccel}. \emph{Virtual touch} computes contact signals from kinematics and motor currents instead of rendering sensor images, which enabled zero-shot force-based grasping and reorientation~\cite{forcegrasp,tac_forcereal}. Finally, some work \emph{avoids tactile simulation altogether}: PTLD collects real data with privileged sensors and distills a tactile state estimator, improving an in-hand rotation benchmark by 182\% over a proprioception-only policy~\cite{tac_ptld}. Results remain sobering for hands: a tactile-only LEAP Hand grasping policy trained via real-to-sim calibration and RL achieved 27\% real success over 20 objects~\cite{tac_blindgrasp}. Unified simulated tactile interfaces across 12 hands~\cite{bench2dex} are a step towards comparable evaluation, but their authors explicitly do not assume simulated touch can replace real touch.

\subsection{Tactile representations and foundation models}
\label{sec:touch-repr}

Tactile representation learning followed the trajectory of vision, one step behind. Paired in-the-wild visuo-tactile data collected by humans enabled the first self-supervised visuo-tactile features~\cite{tac_touchandgo}. UniTouch aligned touch embeddings with pretrained image embeddings, and thereby with language and sound, using sensor-specific tokens~\cite{tac_unitouch}. T3 introduced a shared trunk with sensor-specific encoders, pretrained on over 3M samples from 13 sensors, and improved sub-millimeter multi-pin insertion by 25\% over training from scratch and 53\% over policies without touch~\cite{tac_t3}. Sparsh pretrained self-supervised encoders on more than 460k tactile images and outperformed task- and sensor-specific models by 95.1\% on average on the six-task TacBench~\cite{tac_sparsh}; Sparsh-X extended this to image, audio, motion and pressure from Digit~360 and raised policy success by 63\%~\cite{tac_sparshx}. At policy scale, FTP-1 pretrains on about 3{,}000 hours from 21 sensors and reports gains over baselines of 17.2\% on seen and 31\% on two unseen sensor setups~\cite{tac_ftp1}. Full-hand benchmarks pair egocentric vision with millions of tactile frames over hundreds of tasks~\cite{tac_htbench}. Together with canonical surface representations~\cite{unitachand} and unpaired alignment~\cite{tactalign,anytouch} (\cref{sec:ce-tactile}), these results suggest that cross-sensor transfer is feasible but far from solved.

\subsection{Touch in perception and control}
\label{sec:touch-policy}

\paragraph{State estimation.} The most mature use of touch is perception. NeuralFeels fuses vision and touch on a multi-fingered hand to reconstruct and track objects online with neural fields, reaching 4.7\,mm average pose drift and up to 94\% better tracking than vision alone under heavy occlusion~\cite{tac_neuralfeels}.

\paragraph{RL and imitation.} Touch enables multi-axis and orientation-invariant in-hand rotation~\cite{sim_rotateit,sim_synesthesia,sim_anyrotate,tac_touchdex}. In imitation learning, 3D-ViTac fuses dense flexible tactile pads (3\,mm$^2$ per unit) with vision in a 3D representation and outperforms vision-only diffusion policies on fragile and in-hand tasks~\cite{tac_3dvitac}; Reactive Diffusion Policy pairs a slow latent diffusion policy with a fast tactile tokenizer for high-rate closed-loop control~\cite{tac_rdp}. The \emph{slow--fast split}, also found in T-Rex~\cite{trex} and in industrial contact layers (\cref{sec:learn-hier}), recurs because high-rate touch disrupts slow pretrained models.

\paragraph{Tactile VLAs.} Vision--tactile--language--action models multiplied in 2025--2026. VTLA fine-tunes with preference optimization for fingertip insertion and exceeds 90\% on unseen pegs~\cite{tac_vtla}; Tactile-VLA combines touch with hybrid position--force control and argues that VLM priors already encode physical semantics~\cite{tac_tactilevla}; ForceVLA fuses six-axis force through a mixture of experts and gains 23.2\% over $\pi_0$-based baselines on contact-rich tasks~\cite{tac_forcevla}; OmniVTLA's dual-path tactile encoder reaches 100\% pick-and-place success on dexterous hands (+6.2\% over the baseline) with a 135k-sample tri-modal dataset~\cite{tac_omnivtla}; TacVLA activates tactile tokens only on contact, reporting a 60\% improvement over baselines on in-box picking (the summarized source does not specify whether this is relative or percentage-point improvement) and 2.1$\times$ robustness under occlusion~\cite{tac_tacvla}; and UniTacVLA predicts future touch inside the VLA with a tactile chain of thought~\cite{tac_unitacvla}. Two recent models target dexterous scale: STAR pretrains on about 200 hours of bimanual data (69.5\% of tasks multi-fingered) and reaches 61\% on four real tasks with 100 post-training trajectories each~\cite{tac_star}, and N0-VTLA pretrains latent tactile tokens at scale and adds advantage-conditioned offline RL, reaching 63.8\% on a 20-task simulation suite versus 44.0\% for the strongest baseline~\cite{tac_n0vtla}. Hierarchical tactile systems add tactile sub-goals, residual refiners and tactile world models~\cite{tac_touchworld}.

\paragraph{Touch from humans and in world models.} Human tactile demonstrations scale: UVTA records images, touch and hand motion with tactile motion capture and predicts both future actions and touch, reaching 70\% average success versus 29\% for the strongest visuo-tactile baseline, with no saturation up to 1{,}000 human demonstrations per task~\cite{tac_uvta}. Injecting glove touch into a video diffusion transformer reduces underestimation of hand motion in egocentric prediction from 23\% to 9\%, and helps even when touch is absent at inference~\cite{tac_dtwm}. Handheld UMI-style devices with fingertip touch report corpora above 10k hours, though with gripper-like end effectors~\cite{tac_tacclaw}.

\begin{table}[t]
\centering\tabsize
\caption{Tactile representation and policy models with cross-sensor or dexterous evidence (as reported).}
\label{tab:tactilemodels}
\begin{tabularx}{\columnwidth}{L{1.8cm}L{0.65cm}Y}
\toprule
Model & Year & Pretraining scale / headline result \\
\midrule
T3~\cite{tac_t3} & 2024 & 3M+ samples, 13 sensors; +25\% insertion vs.\ scratch \\
Sparsh~\cite{tac_sparsh} & 2024 & 460k+ images; +95.1\% avg.\ on TacBench \\
Sparsh-skin~\cite{tac_sparshskin} & 2025 & Full-hand magnetic skin; $>$41\% over prior \\
Sparsh-X~\cite{tac_sparshx} & 2025 & $\sim$1M interactions, 4 modalities; +63\% policy \\
OmniVTLA~\cite{tac_omnivtla} & 2025 & 135k samples; 100\% pick-place on hands \\
FTP-1~\cite{tac_ftp1} & 2026 & $\sim$3{,}000\,h, 21 sensors; +31\% unseen sensors \\
STAR~\cite{tac_star} & 2026 & $\sim$200\,h bimanual; 61\% on 4 real tasks \\
N0-VTLA~\cite{tac_n0vtla} & 2026 & Large tactile pretraining + offline RL; 63.8\% (20-task sim.) \\
UVTA~\cite{tac_uvta} & 2026 & 1k human demos/task; 70\% vs.\ 29\% \\
\bottomrule
\end{tabularx}
\end{table}

\takeaway{Touch has moved from sensor papers to pretrained encoders and tactile VLAs, but three obstacles remain: sensor heterogeneity (cross-sensor gains are real but partial), tactile simulation fidelity (sim-to-real touch for hands is still weak), and thin evidence on 20+-DoF hands, since most tactile VLAs are validated on insertion or gripper tasks. Slow--fast designs and contact-gated tactile tokens are the recurring architectural answer.}

\section{Layer 4: Learning and Foundation Models}
\label{sec:learning}

\subsection{Simulation infrastructure}
\label{sec:learn-sim}

GPU-parallel simulation made large-scale RL for hands practical. Isaac Gym keeps physics and policy training on one GPU and reports a speed-up of two to three orders of magnitude over CPU-simulation pipelines~\cite{sim_isaacgym}; Orbit turned it into a modular framework with reusable robots, sensors and actuators~\cite{sim_orbit}; and Isaac Lab, its successor, adds GPU rendering, actuator and sensor models, data collection and domain randomization, with application areas including whole-body control, manipulation and learning from human demonstrations~\cite{sim_isaaclab}. The JAX lineage offers an open alternative: Brax provides massively parallel differentiable rigid-body simulation~\cite{sim_brax}, and MuJoCo Playground trains sim-to-real policies in minutes on a single GPU~\cite{sim_mjplayground}. The SAPIEN/ManiSkill line scales object diversity, with ManiSkill2 providing 20 task families, more than 2{,}000 objects and over 4M demonstration frames~\cite{sim_maniskill2,maniskill3}, while RoboVerse wraps multiple simulators behind one interface~\cite{sim_roboverse}. Two gaps remain. First, none of these platforms benchmarks contact fidelity for dexterous hands directly, and tendon-driven or underactuated hands with sheath friction and joint coupling are hard to model; learned actuator models, introduced for legged robots~\cite{sim_actuatornet}, and joint-wise neural dynamics~\cite{dexndm} partly compensate. Second, tactile simulation lags behind rigid-body simulation (\cref{sec:learn-vtla}).

\subsection{Reinforcement learning and sim-to-real}
\label{sec:learn-rl}

RL in simulation remains the workhorse for in-hand reorientation, hand--arm grasping and force-controlled skills, where demonstrations are hard to provide.

\paragraph{In-hand reorientation.} The lineage starts with OpenAI's Shadow Hand reorientation and Rubik's cube work using massive domain randomization~\cite{dactyl,rubik}, continues with DeXtreme's GPU-parallel training and robust pose estimation on the Allegro Hand~\cite{dextreme} and Visual Dexterity's point-cloud teacher--student distillation for downward-facing reorientation of novel shapes~\cite{visualdexterity}. HORA showed that proprioception alone suffices for z-axis rotation: an adaptive policy trained only on simulated cylinders transfers without fine-tuning and rotates dozens of real objects by inferring object properties from proprioceptive history~\cite{sim_hora}. Later work added touch and multiple axes: RotateIt distills a privileged oracle into a visuotactile transformer~\cite{sim_rotateit}; Robot Synesthesia merges touch and vision in a point-cloud representation~\cite{sim_synesthesia}; AnyRotate performs gravity-invariant multi-axis rotation with dense simulated touch and zero-shot transfer~\cite{sim_anyrotate}. Dynamic skills benefit from combining simulation with small real data: pen spinning pretrains on oracle-generated simulated trajectories and fine-tunes on fewer than 50 real trajectories to spin more than ten pen-like objects~\cite{sim_penspin}. More recent work closes the reality gap systematically with joint-wise neural dynamics learned from autonomously collected real data~\cite{dexndm}, domain randomization in a 3D Gaussian representation for monocular RGB reorientation under hard lighting~\cite{viserdex}, and fingertip-only reorientation on hydraulic hands~\cite{sanctuaryhand}.

\paragraph{Hand--arm grasping, bimanual skills and tool use.} DexPBT uses decentralized population-based training for one- and two-arm hand systems~\cite{sim_dexpbt}; DextrAH-G distills a teacher acting through geometric fabrics, a structured and safe action space, into a depth-based student for fast grasping and transport with a 23-motor arm--hand system~\cite{sim_dextrahg}, and DextrAH-RGB achieves competitive transfer from RGB alone~\cite{sim_dextrahrgb}. Bimanual sim-to-real RL now twists lids off unseen bottles~\cite{sim_twistlids} and performs humanoid bimanual tasks with automatic real-to-sim tuning, generalizable reward design and divide-and-conquer distillation~\cite{humanoidsim2real}; curricula with decaying virtual object controllers handle bimanual articulated objects~\cite{dexmachina,sim_artigrasp}. Tool use combines simulation with real refinement: articulated tools such as scissors, pliers and staplers are handled by a simulation-trained base policy refined on hardware~\cite{sim_atar} or by coarse-to-fine motion synthesis~\cite{mana}, and contact-rich tasks such as nut fastening and screwdriving use simplified-object RL in simulation as a prior for teleoperated tactile demonstrations~\cite{sim_imperfectsim}. Fast tactile simulation with current-to-torque calibration achieved the first zero-shot sim-to-real controllable grip-force tracking~\cite{forcegrasp}; skill chaining extends horizons~\cite{seqdex}.

\paragraph{Automating sim-to-real.} Teacher--student distillation with privileged information is nearly universal for hands~\cite{sim_hora,sim_rotateit,sim_dextrahg,sim_penspin,sim_unigrasptransformer}. LLMs now write reward functions: Eureka's evolutionary reward search outperforms expert-written rewards on 83\% of 29 environments, with an average normalized improvement of 52\%, and enables simulated Shadow Hand pen spinning~\cite{sim_eureka}; DrEureka additionally designs domain-randomization distributions and matches human-designed configurations for dexterous manipulation~\cite{sim_dreureka}. Active exploration collects informative real data for system identification~\cite{sim_asid}. RL also produces data and controllers for other layers: sampling in simulation for retargeting~\cite{spider}, amplification of a few exoskeleton demonstrations~\cite{exostart}, and low-level controllers for teleoperation~\cite{dexteritygen}.

\subsection{Grasp synthesis and grasp policies}
\label{sec:learn-grasp}

Grasping is the most mature dexterous skill, and its data scale has grown from millions to hundreds of millions of synthetic grasps (\cref{tab:grasp}). Three paradigms coexist. \emph{Optimization-based synthesis} directly searches for physically valid grasps: DexGraspNet generated about 1.32M grasps with differentiable force closure~\cite{dexgraspnet}; differentiable grasp-wrench-boundary estimation aligns the grasp wrench space with a task wrench space and is 50$\times$ faster than DexGraspNet at comparable quality~\cite{sim_taskgwb}; BODex uses bilevel optimization with realistic contact models to synthesize more than 49 grasps per second on one GPU, with above 75\% simulated success and below 1\,mm penetration on three hands and 81\% on a real Shadow Hand~\cite{sim_bodex}; and SpringGrasp plans compliant grasps under shape uncertainty~\cite{sim_springgrasp}. \emph{Learned generative samplers} amortize this search: early samplers generated grasps from RGB-D in about a second~\cite{sim_multifingan,sim_shapecompletion}; DexDiffuser combines a diffusion sampler with an evaluator, gaining 9.1\% in simulation and 19.4\% in the real world over a prior generative baseline~\cite{sim_dexdiffuser}; DexGrasp Anything injects physics constraints into diffusion training and sampling and releases more than 3.4M poses on 15k objects~\cite{sim_dexgraspanything}; Get a Grip trains vision-based evaluators on 3.5M grasps over 4.3k objects and shows that performance degrades as the dataset shrinks~\cite{sim_getagrip}; and DexGraspNet~2.0 learns a two-stage diffusion model from 427M grasps in 8{,}270 cluttered scenes, reaching 90.7\% real success in clutter with zero-shot transfer~\cite{sim_dexgraspnet2}. \emph{RL policies distilled from specialists} close the loop: UniDexGrasp++ reaches 85.4\% and 78.2\% success on train and test objects~\cite{sim_unidexgraspplus,unidexgrasp}; UniGraspTransformer distills thousands of per-object policies into a transformer and improves on completely unseen objects by 10.1\% over UniDexGrasp++~\cite{sim_unigrasptransformer}; and RobustDexGrasp reports 97.0\% success over 247{,}786 simulated objects and 94.6\% over 512 real objects with disturbance-resistant dynamic grasping~\cite{sim_robustdexgrasp}.

Grasping is also becoming \emph{task- and taxonomy-aware}. Dexonomy synthesizes 31 GRASP-taxonomy grasp types~\cite{feix} from one annotated template per hand and type, producing 9.5M grasps over 10.7k objects and 82.3\% real success for a type-conditional model~\cite{sim_dexonomy}; AffordDexGrasp performs open-set language-guided grasping with affordance and grasp flow matching~\cite{sim_afforddexgrasp}; and hand-agnostic synthesis through contact maps~\cite{sim_gendexgrasp} and test-time adaptation of human grasps~\cite{sim_grasptta} connect grasping to cross-embodiment transfer (\cref{sec:crossemb}). Because simulators, friction and penetration settings differ, success rates across these papers are not comparable; a shared real-world grasping protocol with standard object sets remains missing.

\begin{table}[t]
\centering\tabsize
\caption{Scale of dexterous grasp datasets and synthesis methods (as reported).}
\label{tab:grasp}
\begin{tabularx}{\columnwidth}{L{2.25cm}L{0.7cm}Y}
\toprule
Work & Year & Scale / headline result \\
\midrule
DexGraspNet~\cite{dexgraspnet} & 2022 & $\sim$1.32M grasps, 5k+ objects (Shadow) \\
UniDexGrasp++~\cite{sim_unidexgraspplus} & 2023 & 85.4\% train / 78.2\% test success \\
Get a Grip~\cite{sim_getagrip} & 2024 & 3.5M grasps, 4.3k objects; learned evaluators \\
DexGraspNet 2.0~\cite{sim_dexgraspnet2} & 2024 & 427M grasps, 8{,}270 scenes; 90.7\% real in clutter \\
BODex~\cite{sim_bodex} & 2024 & $>$49 grasps/s/GPU; 81\% on real Shadow Hand \\
DexGrasp Anything~\cite{sim_dexgraspanything} & 2025 & 3.4M+ poses, 15k+ objects \\
Dexonomy~\cite{sim_dexonomy} & 2025 & 9.5M grasps, 31 GRASP types; 82.3\% real \\
RobustDexGrasp~\cite{sim_robustdexgrasp} & 2025 & 97.0\% on 247k sim objects; 94.6\% on 512 real \\
DexGrasp-Zero~\cite{dexgraspzero} & 2026 & Cross-hand zero-shot: 82\% on 3 real hands \\
\bottomrule
\end{tabularx}
\end{table}

\subsection{Imitation learning}
\label{sec:learn-il}

Behavior cloning with expressive action models made demonstration-based dexterous learning practical: action chunking transformers~\cite{act}, diffusion policies in image~\cite{diffusionpolicy} and point-cloud space~\cite{dp3}, and hierarchical VLM-plus-diffusion systems such as DexGraspVLA, which grasps in thousands of unseen cluttered scenes at above 90\% success~\cite{dexgraspvla}. Typical budgets are a few hundred demonstrations per task~\cite{dexwild,dexumi}.

Several architectural refinements matter specifically for high-DoF hands. \emph{3D representations} help with occlusion and viewpoint change: the improved 3D diffusion policy iDP3 controls a 25-DoF humanoid upper body from egocentric LiDAR point clouds, was evaluated in more than 2{,}000 real rollouts, and generalizes from single-scene training data to new settings with onboard compute only~\cite{pol_idp3}. \emph{Symmetry} improves data efficiency: an SO(2)-equivariant diffusion policy improves average success by 21.9\% over Diffusion Policy on 12 MimicGen tasks~\cite{pol_equidp}, but we found no equivariant policy designed for multi-fingered kinematics. \emph{Fast inference} becomes critical as action dimension and control rate grow: Consistency Policy distills a diffusion policy into a few-step model that is about an order of magnitude faster~\cite{pol_consistencypolicy}, and the FAST tokenizer compresses high-frequency action chunks so that autoregressive VLAs train up to 5$\times$ faster than the $\pi_0$ setup~\cite{pol_fast}. \emph{Fine-tuning recipes} also matter: OpenVLA-OFT combines parallel decoding, action chunking and continuous actions to raise LIBERO success from 76.5\% to 97.1\% with 26$\times$ higher throughput~\cite{pol_oft}. A different route uses \emph{tracking controllers}: DexTrack learns a neural controller that tracks human hand--object references, improving success by more than 10\% over baselines in simulation and on a real hand~\cite{pol_dextrack}. Imitation alone, however, inherits the demonstrations' ceiling, which motivates the post-training methods below.

\subsection{Vision-language-action foundation models}
\label{sec:learn-vla}

Generalist VLAs pair a pretrained VLM with an action decoder and are trained on multi-embodiment data~\cite{rt2,openvla,openx,droid,rdt}. Flow-matching action experts~\cite{pi0} and co-training with web and multi-source data~\cite{pi05} improved open-world generalization; $\pi_{0.7}$ conditions on subtasks, metadata, control modality and visual subgoals to absorb heterogeneous and even failed data, and transfers laundry folding zero-shot to a new robot platform~\cite{pi07}. GR00T N1~\cite{gr00tn1} and N1.7~\cite{gr00tn17} add a ``data pyramid'' of web video, synthetic and real data and, in N1.7, 20k hours of human egocentric video (EgoScale), with Apache-2.0 code~\cite{gr00tgithub}; Gemini Robotics~\cite{geminirobotics,geminirobotics15} and Gemini Robotics~2~\cite{geminirobotics2} extend to whole-body and multi-finger control. Dexterity-specific systems include ByteDance's GR-3~\cite{gr3} and GR-Dexter, which couples a 21-DoF hand, hundreds of hours of egocentric video and cross-embodiment data~\cite{grdexter}, as well as X Square Robot's WALL-OSS~\cite{walloss} and Being-H0/H0.5~\cite{beingh0,beingh05}. Architecturally, most VLAs used with hands follow the ``VLM plus diffusion action head'' pattern popularized by CogACT, which decouples a 7B cognition module from a diffusion action transformer and outperforms OpenVLA by more than 35\% in simulation and 55\% in real experiments~\cite{pol_cogact}. DexVLA scales a plug-in diffusion expert to 1B parameters and trains it in three stages (cross-embodiment expert pretraining, embodiment alignment, task post-training); it is one of the few VLAs evaluated explicitly on a dexterous hand~\cite{pol_dexvla}. Latent-action pretraining from video offers a cheaper route: UniVLA learns task-centric latent actions in DINO feature space and outperforms OpenVLA with less than 1/20 of the pretraining compute and 1/10 of the downstream data~\cite{pol_univla}; Humanoid-VLA uses human motion data and unlabeled video for whole-body control~\cite{pol_humanoidvla}. A caveat comes from benchmarks: DexVerse finds that web-scale pretraining gives no generalization advantage for high-DoF hand control~\cite{dexverse} (\cref{sec:eval}). Most public VLA results are on grippers, and dexterous evidence is still thin.

\subsection{Fast--slow hierarchies with a contact layer}
\label{sec:learn-hier}

\begin{figure*}[t]
\centering
\includegraphics[width=\textwidth]{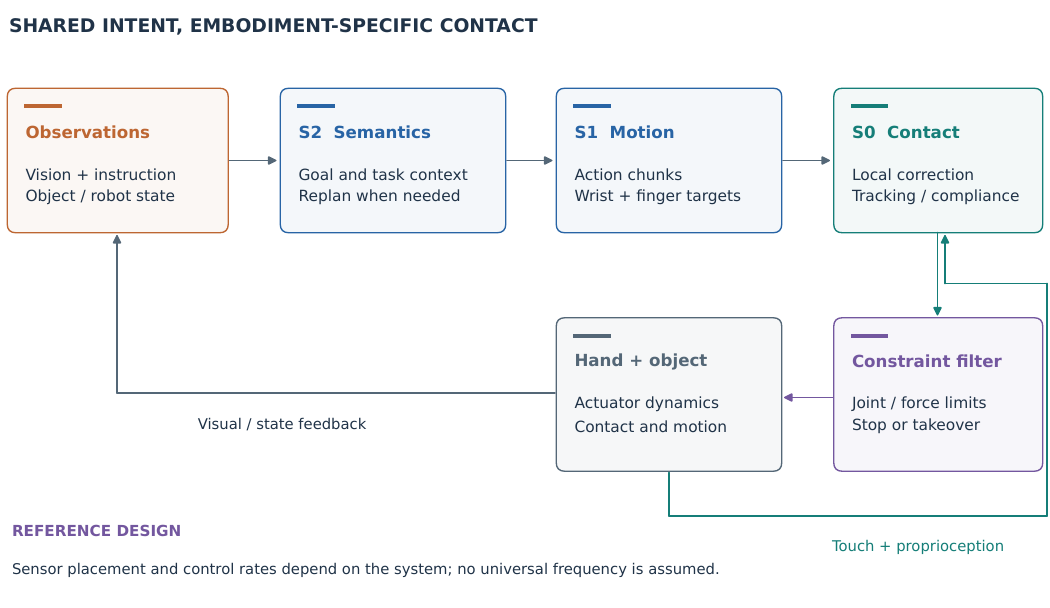}
\caption{Reference architecture separating semantic intent, chunked motion and local contact response, motivated by the hierarchies in \cref{tab:hierarchy}~\cite{figurehelix02,craftnet,agilinkduet,trex}. The constraint filter is a proposed deployment component, not a verified feature of all cited systems. The fast feedback path uses touch and proprioception; visual/state feedback supports replanning. S0 can include whole-body tracking and is not synonymous with a tactile controller.}
\label{fig:control-architecture}
\end{figure*}

\Cref{fig:control-architecture} separates an architectural hypothesis from the results of named systems. Its practical question is whether contact feedback should enter the main policy, a fast local controller, or both. A fair ablation must hold sensing, compute and delay fixed; nominal controller frequency alone does not measure end-to-end reaction latency.

A single network running at one frequency struggles to both ``decide what to do'' and ``modulate force at the moment of contact''. A clear 2026 trend is a hierarchy in which semantic planning is slow, trajectory generation runs at an intermediate rate and contact reaction is fast (\cref{tab:hierarchy}). Figure's Helix~02 uses a semantic S2, an S1 transformer outputting whole-body joint targets including every finger at 200\,Hz, and a 10M-parameter learned S0 whole-body controller at 1\,kHz trained on more than 1{,}000 hours of retargeted human motion plus simulated RL~\cite{figurehelix02}. Sharpa's CraftNet splits a Motion Brain (S1) from an Interaction Brain (S0) that governs contact with objects~\cite{craftnet}. AGILINK's DUET separates a manipulation layer from a contact layer driven by real-time touch on 16--20-DoF hands~\cite{agilinkduet}; AgiBot's GO-2 uses an action chain-of-thought and asynchronous dual systems~\cite{agibotgo2}. Hierarchy also appears in the time domain in T-Rex, whose mixture-of-transformers combines a low-rate action expert with a high-rate tactile expert and improves 12 contact-rich tasks by more than 30\%~\cite{trex}. These industrial designs have academic precursors without touch: HiRT runs a slow VLM and conditions a fast visuomotor policy on its latents, doubling control frequency on static tasks at comparable success and raising success on novel dynamic tasks from 48\% to 75\%~\cite{pol_hirt}; RoboDual pairs a generalist VLA with a 20M-parameter specialist trained on 5\% of the demonstrations, gaining 26.7\% in the real world and running 3.8$\times$ faster~\cite{pol_robodual}; Hi Robot adds a language-reasoning high level that incorporates user feedback~\cite{pol_hirobot}; and Hume lets System~2 sample candidate actions and pick one with a learned value head while System~1 reacts quickly~\cite{pol_hume}. RoboDual's numbers quantify the case for small, fast, specialized layers that the contact layer in dexterous systems takes further. The appeal for hand makers is clear: contact control depends on hardware dynamics and sensor timing, making its interface to general foundation models an important co-design question.

\begin{table}[t]
\centering\tabsize
\caption{Hierarchical control in dexterous systems (company-reported).}
\label{tab:hierarchy}
\begin{tabularx}{\columnwidth}{L{1.6cm}L{1.55cm}L{1.6cm}Y}
\toprule
System & Slow (semantics) & Mid (visuomotor) & Fast (contact/whole body) \\
\midrule
Helix 02~\cite{figurehelix02} & S2 VLM latents & S1, 200\,Hz, fingers incl. & S0, 1\,kHz, 10M params \\
CraftNet~\cite{craftnet} & language/task & S1 Motion Brain & S0 Interaction Brain (contact) \\
DUET~\cite{agilinkduet} & \multicolumn{2}{l}{manipulation layer} & contact layer (touch) \\
GO-2~\cite{agibotgo2} & action CoT & \multicolumn{2}{l}{async. slow intent / fast tracking} \\
T-Rex~\cite{trex} & -- & low-rate action expert & high-rate tactile expert \\
\bottomrule
\end{tabularx}
\end{table}

\subsection{World-action models}
\label{sec:learn-wam}

World models entered the policy backbone in 2026. We distinguish (i)~models that \emph{jointly predict future observations and actions} and act directly, (ii)~models that predict in a \emph{latent} space only, and (iii)~world models used as \emph{evaluators or RL environments}. In group~(i), DreamZero is a 14B autoregressive video-diffusion world-action model running closed-loop at 7\,Hz~\cite{dreamzero}; Liber-0 uses a mixture-of-transformers~\cite{mot} with language, vision and action experts, an embodiment-as-context representation and a first pretraining stage of pure video prediction~\cite{liber0}; PAD jointly denoises future images and actions~\cite{pad}; LDA-1B models dynamics in a DINO latent space~\cite{lda1b}; and industrial programs report similar designs~\cite{psir2,wallb,galbotastra,robotera}. 1X generates future video with a text-conditioned diffusion model and extracts actions with inverse dynamics~\cite{onexredwood}. This group has a longer academic lineage: UniPi framed policies as text-conditioned video generation followed by inverse dynamics~\cite{pol_unipi}; the Unified Video Action model learns a joint video--action latent with separate lightweight heads so that actions can be decoded without generating video~\cite{pol_uva}; WorldVLA unifies image and action generation in one autoregressive model~\cite{pol_worldvla}; and large video generators can be adapted into policies with little demonstration data~\cite{pol_videopolicy}. Instead of acting through generated pixels, Video Prediction Policy feeds predictive representations of a fine-tuned video diffusion model to the policy and reports 31.6\% higher success on complex real-world dexterous tasks~\cite{pol_vpp}. World models are also used to \emph{synthesize data}: DreamGen fine-tunes a video world model, generates new videos, recovers pseudo-actions with a latent action model or inverse dynamics, and enables a humanoid to perform 22 new behaviors from teleoperation data of a single pick-and-place task~\cite{pol_dreamgen}, on top of open world-foundation-model platforms~\cite{pol_cosmos}. In group~(ii), Being-H0.7 inserts learnable latent queries between perception and action, aligned during training with a posterior branch that sees the future and discarded at inference, keeping latency in the few-millisecond-per-step range~\cite{beingh07}; DexWM plans with MPC in a latent world model pretrained on human video~\cite{dexwm}, and V-JEPA~2 pretrains an action-free joint-embedding predictor on more than 1M hours of video and post-trains an action-conditioned variant on less than 62 hours of robot video for zero-shot planning on arms~\cite{pol_vjepa2}. In group~(iii), Genie Envisioner~\cite{genieenvisioner} and 1X's world model~\cite{onexredwood} are used to rank policy versions and run imagined RL. The open questions are inference cost, how to evaluate physical consistency, and whether pixel-level prediction is necessary for contact-rich hand control.

\subsection{Touch inside the model: VTLA}
\label{sec:learn-vtla}

Vision--tactile--language--action (VTLA) models treat touch as a first-class input. CraftNet optimizes its training data by mapping tactile signals onto human video and glove data~\cite{craftnet}; Figure~03 feeds palm cameras and fingertip force into S1~\cite{figure03,figurehelix02}; TouchScale shows that visuo-tactile mid-training on human glove data improves real contact tasks~\cite{touchscale}; and tactile kinesthetic teaching provides contact-aware supervision~\cite{dextac}. Tactile simulation fidelity and sensor heterogeneity are the main barriers; \cref{sec:touch} reviews tactile VLAs and the supporting sensors, simulators and representations in detail.

\subsection{RL post-training, self-improvement and prompting}
\label{sec:learn-post}

After imitation, policies improve through RL and failure recycling. Three recipes are converging. (i)~\emph{Online policy-gradient RL on the VLA itself}, enabled by vectorized simulation and new infrastructure: VLA-RL~\cite{pol_vlarl}, SimpleVLA-RL, which reports emergent behaviors absent from the demonstrations~\cite{pol_simplevlarl}, RLinf-VLA, with 98.1\% on 130 LIBERO tasks~\cite{pol_rlinf}, and $\pi_{\text{RL}}$, which makes flow-matching VLAs trainable with RL~\cite{pol_pirl}. (ii)~\emph{Diffusion-specific RL}: policy-gradient fine-tuning of diffusion policies~\cite{pol_dppo} and steering a frozen policy by RL in its latent-noise space~\cite{pol_dsrl}. (iii)~\emph{Residual refinement on a frozen base}: ResiP adds a closed-loop residual RL policy to a chunked imitation policy for precise assembly~\cite{pol_resip}, and Policy Decorator refines large imitation policies online, including on the Adroit dexterous-hand tasks~\cite{pol_policydecorator}. In the real world, human-in-the-loop RL trains vision-based policies in 1--2.5 hours~\cite{pol_hilserl}, and ConRFT reaches 96.3\% average success on eight real tasks after 45--90 minutes of online fine-tuning of a VLA with a consistency head~\cite{pol_conrft}; interactive imitation interfaces let a human take over and hand back control during bimanual tasks~\cite{pol_robocopilot}. Almost all of this evidence is on grippers; scaling real-world RL to 20+-DoF hands, with their wear and safety constraints, is open. ByteDance's GR-RL targets precise long-horizon dexterous manipulation~\cite{grrl}; $\pi_{0.7}$ turns failed autonomous episodes into useful training data through metadata conditioning~\cite{pi07}; and DITTO-X uses reverse teleoperation for human interventions~\cite{dittox}. Generalist's GEN-1.5 shows in-context ``physical prompting'': placing a 3--12\,s demonstration in a 30\,s context lets the model perform a new task without gradient updates, with 59\% average success over ten tasks from one demonstration and 83\% after ten gradient steps on five minutes of data~\cite{gen15}. Its hardware is pincer-style rather than multi-fingered.

\subsection{Whole-body and mobile dexterity}
\label{sec:learn-humanoid}

Dexterous hands are increasingly mounted on humanoids and mobile bases, so hand policies must coexist with whole-body control. The prevailing structure is a general whole-body \emph{tracking} controller trained with RL in simulation, below which hands are often treated as simple end effectors, and above which teleoperation or learned policies issue targets. H2O performs real-time whole-body teleoperation from an RGB camera with zero-shot sim-to-real transfer and filters infeasible retargeted human motion in simulation~\cite{pol_h2o}; HOVER unifies multiple command modes (e.g., root velocity and upper-body joint angles) in one controller~\cite{pol_hover}; ExBody2 uses teacher--student training on filtered motion capture and documents a trade-off between tracking generality and fine-tuned motions~\cite{pol_exbody2}; TWIST trains a single RL-plus-BC controller from motion capture for whole-body teleoperation~\cite{pol_twist}; AMO expands the manipulation workspace of a 29-DoF humanoid with adaptive motion optimization~\cite{pol_amo}; and ResMimic adds a residual policy for object loco-manipulation on top of a tracker pretrained on human motion, using object-tracking and contact rewards and a virtual-object curriculum similar to DexMachina~\cite{pol_resmimic,dexmachina}. Simulation data generation with LLM-reasoned task plans produces bimanual dexterous humanoid demonstrations on which 2D and 3D diffusion policies scale~\cite{pol_humanoidgen}. For mobile bimanual manipulation, Mobile ALOHA showed that 50 demonstrations per task plus co-training with static data raise success by up to 90\%~\cite{pol_mobilealoha}; equivalent results with multi-fingered hands are still scarce, and the interface between whole-body controllers and high-frequency finger contact control is largely unexplored.

\begin{table*}[t]
\centering\tabsize
\caption{Industrial and large-scale programs relevant to dexterous hands (as of November 2026). Results are company-reported and not comparable.}
\label{tab:industry}
\begin{tabularx}{\textwidth}{L{2.4cm}L{2.6cm}L{1.9cm}L{5.0cm}Y}
\toprule
Organization & Model (latest) & Route & Architecture notes & Data / dexterity evidence \\
\midrule
Physical Intelligence & $\pi_0\to\pi_{0.5}\to\pi_{0.7}$~\cite{pi0,pi05,pi07} & VLA, RL & Flow-matching expert; context conditioning incl.\ metadata & Zero-shot cross-embodiment folding (grippers) \\
NVIDIA & GR00T N1$\to$N1.7~\cite{gr00tn1,gr00tn17}; DreamZero~\cite{dreamzero} & VLA, human data, WAM & Relative EEF actions; embodiment adapters & 20k\,h human video; +54\% on 22-DoF hand~\cite{egoscale} \\
Google DeepMind & Gemini Robotics 1.5 / 2~\cite{geminirobotics15,geminirobotics2} & VLA & Thinking VLA; Motion Transfer & Multi-embodiment transfer; multi-finger control \\
Figure AI & Helix 02~\cite{figurehelix02} & Hierarchical & S2/S1 200\,Hz/S0 1\,kHz & Fingers in S1; palm cams, fingertip force~\cite{figure03} \\
Sharpa & CraftNet~\cite{craftnet} & Hierarchical VTLA & Motion Brain + Interaction Brain & 22-DoF tactile hand~\cite{sharpawave}; glove data \\
Generalist AI & GEN-0$\to$GEN-1.5~\cite{gen15} & Human data & 100\,Hz actions, in-context prompting & Hundreds of thousands of hours (press-reported)~\cite{generalistprofile} \\
Sunday Robotics & ACT-1~\cite{sundayact1} & Human data & Skill Transform from glove to robot & $\sim$10M glove episodes~\cite{sunday36kr}; no robot training data \\
1X Technologies & Redwood, 1X WM~\cite{onexredwood} & VLA, WAM & Video diffusion + inverse dynamics & World-model evaluation \\
Skild AI & Skild Brain~\cite{skildbrain} & Omni-bodied & One brain for many bodies & Sim + video pretraining \\
LiberAI & Liber-0~\cite{liber0} & WAM, glove & MoT, embodiment-as-context & 2.2M\,h; 36\% on 60 zero-shot tasks \\
ByteDance Seed & GR-3, GR-RL, GR-Dexter~\cite{gr3,grrl,grdexter} & VLA, RL & Hardware--model--data co-design & 21-DoF hand, 56-DoF bimanual system \\
BeingBeyond & Being-H0$\to$H0.7~\cite{beingh0,beingh05,beingh07} & Human data, WAM & Unified action space, mixture-of-flow & 35k\,h, 30 robots \\
AgiBot & GO-1/GO-2, GE~\cite{agibotworld,agibotgo2,genieenvisioner} & VLA, WAM, RL & Latent actions; async dual system & 1M+ trajectories \\
AGILINK & DUET~\cite{agilinkduet} & Hierarchical, touch & Manipulation + contact layer & 16--20-DoF hands \\
PsiBot, X Square, Galbot, Robot Era & Psi-R2, WALL-B, AstraBrain, ERA-42~\cite{psir2,wallb,walloss,galbotastra,robotera} & WAM & World model fused into the policy & Human-data or simulation-heavy pipelines \\
\bottomrule
\end{tabularx}
\end{table*}

\takeaway{For high-DoF contact-rich tasks, recent results motivate adding touch and testing fast local feedback. Neither the necessity of touch for every task nor the universal superiority of a separate contact layer has been established. Human data provide scale if hardware is matched or the conversion is reliable; RL and failure recycling push beyond the demonstration ceiling; world models are increasingly used to evaluate and accelerate training, while their role inside the real-time control loop is still debated.}

\section{Layer 5: Evaluation and Applications}
\label{sec:eval}

\subsection{A skill taxonomy}
\label{sec:eval-skills}

We group dexterous behaviors into eight atomic skills, ordered roughly by maturity: \emph{grasping} (stable and functional grasps of arbitrary objects; deployed in logistics and machine tending~\cite{dexgraspnet,unidexgrasp,dexgraspvla}); \emph{in-hand reorientation} (lab-mature through RL~\cite{dactyl,dextreme,visualdexterity,dexndm,viserdex} but with little direct industrial demand); \emph{tool use} (hammers, tweezers, syringes, power tools~\cite{mana,dexjoco,bench2dex,handson}); \emph{bimanual coordination}~\cite{bidexhands,arctic,oakink2,gigahands,bidex100}; \emph{articulated objects} such as lids, doors and scissors~\cite{dexart,dexmachina}; \emph{deformable and fragile objects} such as cables, cloth, eggs and pills, which need touch and fast reaction~\cite{trex,handson}; \emph{insertion and precision assembly} with sub-millimeter tolerance~\cite{trex,dexverse}; and \emph{force-controlled dosing} such as powder weighing, pipetting and writing~\cite{labimus,labdex,forcegrasp}. A grasp taxonomy~\cite{feix} and the Kapandji opposition test remain the standard descriptors of hand capability.

\subsection{Benchmarks and datasets}
\label{sec:eval-bench}

\Cref{tab:bench} lists benchmarks and human--object datasets. Before 2025, benchmarks focused on grasping, reorientation and bimanual RL in simulation~\cite{bidexhands,dexgraspnet,robopianist,dexart,maniskill3}, while human--object datasets provided motion priors~\cite{arctic,oakink2,gigahands}. TeleOpBench compared four teleoperation modalities in mirrored simulation and real settings~\cite{teleopbench}. Between May and October 2026 at least nine new dexterous benchmarks appeared. Four patterns stand out.

\paragraph{From ``pick it up'' to functional tasks.} DexJoCo focuses on tasks that only a dexterous hand can do, such as tool use, reasoning, bimanual coordination and long horizons, with 1.1k demonstration trajectories~\cite{dexjoco}; DexVerse defines 100 tasks decoupled from embodiment~\cite{dexverse}; Bench2Dex offers 26 bimanual tasks on 12 hands with a unified simulated tactile interface~\cite{bench2dex}; BiDex-100 and HandsOn (18k demonstrations across 5 arms and 6 hands) scale task counts further~\cite{bidex100,handson}.

\paragraph{General models are weak at high-DoF control.} DexVerse evaluates DP, DP3, OpenVLA and $\pi_{0.5}$ and finds no generalization advantage from web pretraining for high-DoF hands, with near-universal failure on fine contact tasks~\cite{dexverse}. On HandsOn, a general-purpose frontier model without fine-tuning reaches 5.2\% on demonstration-learning tasks and 1.9\% on transfer tasks~\cite{handson}.

\paragraph{Touch becomes an evaluation axis.} TactiDex moves evaluation from ``moving like a human'' to ``making contact like a human''~\cite{tactidex}; Bench2Dex and T-Rex make touch a first-class signal~\cite{bench2dex,trex}.

\paragraph{Laboratories are a new frontier.} Labimus encodes an organic-chemistry weighing procedure with particle-based powders and instrument read-outs and finds that task completion and experimental validity diverge~\cite{labimus}; LabDex structures real and simulated demonstrations into atomic, composite and long-horizon chemistry tasks~\cite{labdex}; DexBioLab adds flames, vapor and bubbles~\cite{dexbiolab}. Laboratories have standardized procedures, human-designed tools and quantifiable precision, which make them attractive test beds.

\paragraph{Competitions.} The dexterous-hand event of the 2026 World Humanoid Robot Games in Beijing comprised eight five-minute tasks (picking beans with tweezers, opening bottles, weighing 20\,g$\pm$0.5\,g of powder, building blocks, nailing, connecting cables, assembling with a power tool, unboxing), with autonomous runs weighted twice as heavily as teleoperated ones; autonomous solutions proved much harder to qualify~\cite{whrgcomp}. Such scoring rules are beginning to act as de facto acceptance criteria.

\begin{table*}[t]
\centering\tabsize
\caption{Dexterous benchmarks, datasets and competitions.}
\label{tab:bench}
\begin{tabularx}{\textwidth}{L{2.2cm}L{0.95cm}L{1.3cm}L{4.3cm}L{3.3cm}Y}
\toprule
Name & Date & Type & Tasks & Embodiment / platform & Scale / finding \\
\midrule
Bi-DexHands~\cite{bidexhands} & 2022 & Sim & Bimanual skills staged by infant development & Two Shadow hands, Isaac Gym & RL solves tasks at $\sim$48-month level \\
DexGraspNet~\cite{dexgraspnet} & 2022 & Sim data & Grasp synthesis & Shadow Hand & $\sim$1.32M grasps, 5k+ objects \\
ARCTIC~\cite{arctic} & 2023 & Human data & Bimanual articulated objects & MANO + mocap & 2.1M frames \\
RoboPianist~\cite{robopianist} & 2023 & Sim & Bimanual piano playing & Two Shadow hands & High-rate precision \\
DexArt~\cite{dexart} & 2023 & Sim & Faucets, lids, laptops, buckets & Allegro, SAPIEN & Generalization across instances \\
OakInk2~\cite{oakink2} & 2024 & Human data & Affordance--primitive--task hierarchy & Multi-view + mocap & Task-aware motion generation \\
GigaHands~\cite{gigahands} & 2024 & Human data & Everyday bimanual activities & Dense multi-view & 34\,h, 417 objects \\
TeleOpBench~\cite{teleopbench} & 2025 & Sim + real & Bimanual teleoperation & 4 teleop modalities & Sim predicts real ranking \\
DexJoCo~\cite{dexjoco} & 2026-05 & Sim & 11 functional tasks & Franka + Allegro, MuJoCo & 1.1k trajectories; VLAs limited \\
Labimus~\cite{labimus} & 2026-06 & Sim & Chemistry weighing SOP & Humanoid; powder physics & Completion $\neq$ validity \\
T-Rex~\cite{trex} & 2026-06 & Real & 12 contact-rich tasks & Bimanual hands + touch & 100\,h tactile data \\
DexVerse~\cite{dexverse} & 2026-07 & Sim & 100 tasks & Many arm--hand combos & No benefit from web pretraining \\
TactiDex~\cite{tactidex} & 2026-07 & Real & Tactile-guided uni-/bimanual & Full-hand touch & 700+ sequences, 49 objects \\
LabDex~\cite{labdex} & 2026-08 & Real + sim & Atomic$\to$composite$\to$long chemistry & Franka FR3 + XHand & Skill-composition analysis \\
WHRG dexterous event~\cite{whrgcomp} & 2026-08 & Competition & 8 tasks, 5 min each & Any hand; autonomy weighted 2$\times$ & $\sim$51 teams/task \\
Bench2Dex~\cite{bench2dex} & 2026-09 & Sim & 26 bimanual tasks & 12 hands, Isaac Lab, sim touch & Human teleop demos \\
DexBioLab~\cite{dexbiolab} & 2026-10 & Sim & 14 biology-lab tasks & 4 embodiments & Dynamic phenomena \\
BiDex-100~\cite{bidex100} & 2026-10 & Sim & 100+ bimanual tasks & High-DoF hands & Multi-dimensional metrics \\
HandsOn~\cite{handson} & 2026-10 & Sim & Core-50 + Transfer-16 & 5 arms, 6 hands, $\sim$200 assets & 18k demos; 5.2\%/1.9\% zero-shot \\
\bottomrule
\end{tabularx}
\end{table*}

\subsection{Evaluation methodology}
\label{sec:eval-method}

Dexterous papers typically report 10--20 real trials per condition without confidence intervals, and use heterogeneous metrics. Methodology developed for generalist gripper policies can be adopted. \emph{Statistical rigor:} STEP is a sequential test with near-optimal stopping that needs up to 32\% fewer trials than prior procedures while avoiding p-hacking~\cite{dep_step}; anytime-valid sequential comparisons with partial-credit and continuous metrics reduce evaluation burden by up to 70\% relative to batch methods and show that progress metrics separate policies faster than binary success~\cite{dep_beyondbinary}, which supports progress scores for long-horizon dexterous tasks if they are reported alongside success. \emph{Protocol design:} blinded, randomized evaluation in simulation and the real world quantified the benefits of large behavior-model pretraining with statistical confidence~\cite{dep_lbm}; RoboArena distributes double-blind pairwise comparisons across seven institutions and more than 600 real episodes~\cite{dep_roboarena}; and RoboEval adds coordination, safety/stability and stage-wise failure metrics that separate policies with similar success rates~\cite{dep_roboeval}. \emph{Scalable proxies:} SIMPLER showed that simulation aligned in control and appearance correlates strongly with real performance for gripper policies~\cite{dep_simpler}; video-derived digital twins with VLM scoring extend this idea~\cite{dep_robotarenainf}; and AutoEval automates queueing, success detection and resets around the clock~\cite{dep_autoeval}. None of these has yet been validated for in-hand manipulation, where dropped objects require re-staging and contact-rich behavior is hard to simulate faithfully; a dexterous analog of SIMPLER and of automatic reset is an open need (\cref{sec:challenges}).

\subsection{A minimum evaluation record}
\label{sec:eval-record}

The following record is a proposed protocol for future comparisons, not a claim that existing benchmarks already provide all fields. Report each task--hand pair separately before computing an aggregate. Split unseen objects, tasks, environments and embodiments into distinct tests; ``zero-shot'' must state whether target-hand calibration, demonstrations, simulator tuning or interventions were allowed. For binary success, report successful trials and the denominator with an interval and its method. For heterogeneous tasks or repeated trials on the same object, account for clustering rather than treating every attempt as independent. Progress scores require their scoring rule and should remain separate from binary completion.

\begin{table}[t]
\centering\tabsize
\caption{Proposed minimum record for a dexterous evaluation.}
\label{tab:eval-record}
\begin{tabularx}{\columnwidth}{L{1.55cm}Y}
\toprule
Dimension & Record and report \\
\midrule
Embodiment & Active/total DoF; coupling; controller; sensor type, layout and calibration \\
Adaptation & Target-hand data, tuning compute, calibration, interventions and privileged state \\
Generalization & Held-out objects, tasks, sites and hands; possible training overlap \\
Task outcome & Successes/trials; per-task results; progress rubric; uncertainty method \\
Physical outcome & Pose/dosing error, contact-force limits, slip and object damage \\
Timing & End-to-end latency, update rate, chunk horizon and deadline misses \\
Deployment & Attempts, operating hours, takeovers/hour, repairs, downtime and recovery \\
\bottomrule
\end{tabularx}
\end{table}

For deployment, choose acceptance thresholds before testing and tie them to the task's consequences and workload. A short trial with no observed incidents does not establish a zero incident rate. Likewise, a controller running at 1\,kHz can still react slowly if perception and communication dominate the delay. These distinctions prevent nominal rates and attractive headline percentages from substituting for physical reliability.

\subsection{Safety}
\label{sec:eval-safety}

Collaborative-robot safety standards were written for arms. Biomechanical studies behind them show that most ISO/TS~15066 limits are preliminary because they rest on unverified literature data, with pain thresholds measured across 28 body locations in 112 participants~\cite{dep_biomechlimits}; impact thresholds also drop markedly for sharp contact (e.g., about 168\,N for a wedge versus 272\,N for a cylinder at the deltoid, 75th percentile)~\cite{dep_painthreshold}, which matters for fingertips and tools held by hands. We found no standard that addresses pinch and crush hazards specific to multi-fingered hands. On the learning side, constrained RL for VLAs reduced cumulative safety-violation cost by 83.6\% while slightly improving success in mobile manipulation~\cite{dep_safevla}, and dynamic human-to-robot handover with dexterous hands now has a real-world benchmark~\cite{dep_dexh2r}. Practical deployments therefore pair learned policies with deterministic safety layers (speed and force limits, keep-out zones), as in the kitchen design of \cref{sec:case-kitchen}.

\subsection{Applications and deployment maturity}
\label{sec:eval-apps}

\Cref{tab:maturity} gives our assessment of maturity by domain, based on the evidence above. Industrial deployments first appear in \emph{semi-structured fine work}: electronics and automotive part assembly, micro-motor assembly with cycle times of about 9--10\,s, and parcel and pharmaceutical sorting in logistics centers~\cite{mir2026}; humanoids with dexterous hands entered automotive plants for logistics~\cite{figure03}; and foundation-model companies demonstrate part kitting, phone boxing and box folding~\cite{gen15,generalistfunding}. Home tasks such as loading a dishwasher are mostly demonstrations~\cite{figurehelix02}. The best-documented long-duration home studies still use grippers: Dobb-E collected 13 hours in 22 homes and reached 81\% success over 109 tasks in 10 homes during about 30 days~\cite{dep_dobbe}; robot utility models with verifier-triggered retries achieved about 90\% zero-shot success in unseen environments~\cite{dep_rum}; and RL from experience with interventions more than doubled throughput and roughly halved failures on hard real-world tasks such as laundry folding and espresso making~\cite{dep_pistar06}. Mobile bimanual dexterous systems that convert human hand motion with RL are emerging~\cite{dep_hermes}, but without comparable long-duration statistics. In medicine, imitation learning on surgical robots achieved fully autonomous steps of gallbladder removal on eight unseen ex vivo specimens~\cite{dep_srth,dep_srt}, and a teleoperated humanoid with a commercial six-DoF hand performed laparoscopic cholecystectomy on live pigs as a feasibility study~\cite{surgerynature}. We found no peer-reviewed economic or return-on-investment study of dexterous manipulation; deployment economics are documented only in vendor and analyst material. Industry reports suggest that most customers are still at the proof-of-concept stage~\cite{mir2026}.

Field requirements are demanding. Operators in food service, for instance, typically expect more than 95\% task reliability before committing to long-term use, whereas a reported bimanual food-assembly system reached 75\% full-task success after 26 hours of training~\cite{chefrobotics}. For a commercial-kitchen assistant, we propose illustrative acceptance targets of task success above 95\%, at most about one human takeover per operating hour at peak, availability near 98\%, cycle times within 1.2--1.5$\times$ those of a skilled worker, mean time between failures in the hundreds of hours, food-safe washable hand covers, and robustness to steam, grease and wet surfaces. Two architectural lessons follow: dexterous hands should be reserved for steps that truly need them (pairing a tactile 20+-DoF hand with a quick-change tool hand reduces failure points), and quantities that vision estimates poorly, such as dosing liquids or powders, should be closed with additional sensors (e.g., a scale) in the contact layer rather than left to the visuomotor policy.

\begin{table}[t]
\centering\tabsize
\caption{Maturity by skill and domain (our assessment): research activity, benchmark coverage, industrial deployment (3 high, 2 medium, 1 low).}
\label{tab:maturity}
\begin{tabularx}{\columnwidth}{L{2.25cm}cccY}
\toprule
Skill / domain & Res. & Bench. & Deploy & Main bottleneck \\
\midrule
Grasping / sorting & 3 & 3 & 3 & Long tail, transparent objects, cycle time \\
In-hand reorientation & 3 & 2 & 1 & Pose estimation; little demand \\
Tool use & 2 & 3 & 2 & Functional contact choice, impacts \\
Bimanual & 3 & 3 & 2 & Error accumulation, synchronization \\
Deformable / fragile & 2 & 1 & 1 & Simulation gap; needs touch \\
Insertion / assembly & 2 & 1 & 2 & Sub-mm precision, insertion force \\
Laboratory & 2 & 3 & 1 & Completion vs.\ validity \\
Household & 2 & 2 & 1 & Diversity, reliability, safety \\
Medical & 1 & 1 & 1 & Regulation, reliability, haptics \\
\bottomrule
\end{tabularx}
\end{table}

\section{Full-Stack Case Studies}
\label{sec:cases}

The preceding sections analyzed each layer separately. In this section we examine five systems that span several layers, to show how design decisions propagate through the stack and where the evidence is strong or weak.

\subsection{Human video to an open humanoid stack: EgoScale and GR00T}
\label{sec:case-egoscale}

EgoScale~\cite{egoscale} is the clearest public example of a data-first stack. \emph{Data:} 20{,}854 hours of egocentric human video with estimated 21-keypoint hand poses (including EgoDex~\cite{egodex}), plus an aligned set of about 50 hours of human data recorded with gloves, wrist trackers and wrist cameras arranged as on the robot, and about 4 hours of robot data. \emph{Representation:} keypoints are retargeted per frame into the joint space of a 22-DoF Sharpa Wave hand~\cite{sharpawave}; joint-space actions outperform wrist-only or fingertip-only pretraining targets. \emph{Model:} a GR00T-style VLM with a diffusion-transformer action expert and embodiment-specific adapters~\cite{gr00tn1}. \emph{Training:} human pretraining, then human--robot aligned mid-training, then task post-training. \emph{Evidence:} validation loss falls log-linearly with human hours; task completion rises from 0.30 at 1k hours to 0.71 at 20k hours; average success improves by 54\% over no pretraining; and transfer to a three-finger Unitree hand still yields large gains. The corpus was subsequently folded into GR00T N1.7~\cite{gr00tn17}, whose code is released under Apache-2.0 and which uses relative end-effector actions~\cite{gr00tgithub}, and a reference humanoid pairs a Unitree H2-series body with two Sharpa Wave hands~\cite{gr00tref}. \emph{What this case shows:} human data can drive dexterous learning at scale, in a demonstrated setting with a small, carefully aligned bridge dataset and a hand whose kinematics are close to human. \emph{Open questions:} how far the log-linear trend extends, how much of the gain depends on the human-like 22-DoF target, and whether the recipe works equally well for strongly non-anthropomorphic hands.

\subsection{Glove, human-like robot and world-action model: Liber-0}
\label{sec:case-liber}

Liber-0~\cite{liber0} combines three components that each address one component of the human-to-robot gap. For the \emph{supervision gap}, a 131\,g, 0.3\,mm skin-colored glove records measured hand pose together with one egocentric and two wrist cameras, without a robot present. For the \emph{visual gap}, the robot (LEO) has a slim forearm, a three-DoF human-like wrist, an adjustable waist, and chest and wrist cameras mounted where the glove's cameras are; the hand is a commercial WUJI2 hand. For the \emph{embodiment gap}, the model conditions action prediction on a shared encoding of human and robot geometry and kinematics. The model is a world-action model built on a mixture-of-transformers~\cite{mot} with language, vision and action experts; it is pretrained in three stages (video-only prediction, joint world--action pretraining, then manipulation-focused pretraining that upweights glove and teleoperation data) on more than 2.2M hours of human experience, and co-trained on VQA, segmentation, image editing and video understanding.

The value of this report lies in its controlled comparisons. Under identical post-training, 2M hours of web video reach 64.6/51.2 progress on pick-and-place/non-pick-and-place tasks, both below the scores obtained with only 10k hours of glove data; glove plus 25\% of the robot demonstrations reaches 75.6 versus 72.8 for all robot demonstrations; glove data alone (60.3) beats a quarter of the robot data (50.0); skills that exist only in glove data rise from 2.2 to 47.5 progress (54.6 with all glove data), approaching full teleoperation (56.3); more diverse collectors help at fixed volume (12.7$\to$40.1 glove-only); and the human-like robot beats a conventional arm by 28.3 points with the same glove data. Zero-shot success over 60 tasks (600 trials) is 36.0\% overall, but only 10.5\% on multi-step tasks.

Several caveats limit what can be concluded. The report is a blog post without code, weights, parameter count, control frequency or latency; most internal metrics are progress scores, which are more lenient than binary success; per-task trial counts are small (3--10); the zero-shot baseline is a reasoning agent rather than a VLA; manipulation-focused pretraining may overlap with the evaluation task families, which would inflate the claim that post-training adds little (81.4/72.3 before versus 83.6/73.8 after); and the glove carries no touch, which the authors identify as the next step. Genesis AI pursues a similar full-stack route, pairing its own hand with a 1:1 sensorized glove~\cite{genesisai}; together with Liber-0's use of a third-party hand, this suggests an emerging division of labor between model companies and hand makers.

\subsection{Hardware-matched gloves at scale: Sunday and Generalist}
\label{sec:case-gloves}

Two US companies pursue the most aggressive form of robot-free capture. Sunday's Skill Capture Glove shares the geometry and sensor layout of its robot hand, and a Skill Transform pipeline removes human-specific features; the company reports that glove data are converted into robot data with a 90\% success rate and that the resulting ACT-1 model performs long-horizon chores such as clearing a table into a dishwasher without any robot training data~\cite{sundayact1}, while press reports state that the glove sells for about US\$200 and was sent to more than 2{,}000 contributors~\cite{sunday36kr}. Generalist distributes wearable pincer-style ``data hands'' isomorphic to its own end effectors with press reports of more than 270k hours at the time of GEN-0 and more than 500k hours for GEN-1~\cite{generalistprofile}; GEN-1 is pretrained primarily on human-demonstration data and reportedly needs about one hour of robot data per task~\cite{generalistfunding}, and GEN-1.5 introduces in-context learning from 3--12\,s demonstrations~\cite{gen15}. Both cases show that \emph{isomorphism plus volume} can substitute for teleoperation, but neither uses a 20+-DoF multi-fingered hand at scale, and neither has released data or models for independent evaluation.

\subsection{A vertically integrated tactile stack: Sharpa}
\label{sec:case-sharpa}

Sharpa illustrates the hand-maker route into algorithms. The Wave hand has 22 active DoF at human scale with a dynamic tactile array and torque sensing per finger~\cite{sharpawave}; the AE01 glove (22 DoF) provides haptic feedback for training-data collection and remote operation~\cite{sharpaw02}; CraftNet is a hierarchical vision--tactile--language--action model whose Interaction Brain (S0) governs contact with objects while a Motion Brain (S1) directs movement, and which optimizes its training data by mapping tactile signals onto human video and glove data~\cite{craftnet}; Dex-X learns visuo-tactile manipulation from human video with simulated interaction and transfers zero-shot to a 29-DoF hand--arm platform~\cite{dexx}. The company also became the hand supplier of the GR00T reference humanoid (\cref{sec:case-egoscale}). The lesson is that contact control depends on hardware and sensor timing, so hand makers can contribute a distinct co-design point while still using general-purpose foundation models for scale.

\subsection{Designing a deployment: a commercial-kitchen assistant}
\label{sec:case-kitchen}

We close with a deployment design exercise for a wheeled dual-arm robot with dexterous hands in a chain-restaurant kitchen, to illustrate how the requirements of a real setting constrain every layer. Automated stir-fry machines are already deployed at scale in some Chinese restaurant chains~\cite{xiaocaiyuan}, and research on kitchen robots has so far focused on recipe-level planning and dishware perception with grippers~\cite{dep_cookingpddl,dep_kitchenfm}; the open opportunity is the \emph{connecting} work between machines and people. \Cref{tab:kitchen} rates kitchen steps by how much they need a dexterous hand and how hard they are.

\begin{table}[t]
\centering\tabsize
\caption{Commercial-kitchen steps: need for a dexterous hand, difficulty and suggested phase (P1 = first deployment). Illustrative analysis.}
\label{tab:kitchen}
\begin{tabularx}{\columnwidth}{L{1.55cm}L{2.15cm}ccY}
\toprule
Step & Typical actions & Dex. need & Diff. & Phase \\
\midrule
Fetch \& load & Open containers, pour into machine, scrape & Med. & Med. & P1 \\
Seasoning & Scoop powder, squeeze bottles, tear sachets & High & Med.--high & P1 \\
Packing & Box, close lids, add cutlery, bag & High & Med. & P1 \\
Plating & Portion, arrange, wipe rims & High & High & P2 \\
Dish handling & Scrape, sort, rack & Med. & Med. & P2 \\
Cleaning & Wipe, return tools & Med. & Med. & P2 \\
Prep & Peel, pick, wash & High & High & P3 \\
Cutting & Knife work & High & High & Later \\
\bottomrule
\end{tabularx}
\end{table}

\paragraph{Hardware.} A compact base (footprint around 0.6$\times$0.7\,m) with a lift, two 7-DoF arms with at least 5\,kg payload, a 20+-DoF tactile, backdrivable main hand and a low-DoF gripper with quick-change tools (spoon, tongs, scraper) as the second hand. Using the dexterous hand only where it is necessary reduces wear and failure points (\cref{sec:hw-durability}); food-safe, washable silicone covers protect the hand and separate raw from cooked food. Palm cameras compensate for steam and occlusion; a workstation scale closes dosing loops.

\paragraph{Algorithms.} A task agent connected to the point-of-sale and kitchen display systems turns orders into steps; an S2 VLM tracks sub-task progress and counts portions; an S1 VLA produces whole-body action chunks; an S0 contact layer at 0.5--1\,kHz handles grip force, slip and weight-based stopping; and a rule-based safety layer, independent of learning, enforces hot zones, speed limits near people and force limits. World models are used for version ranking and imagined RL, not in the real-time loop (\cref{sec:learn-hier,sec:learn-wam}).

\paragraph{Data.} Cooks wear egocentric glasses and a tactile glove isomorphic to the main hand during normal service; teleoperation is reserved for failure-prone steps; a digital twin of each kitchen supports RL and evaluation; and autonomous episodes, failures and takeovers flow back for targeted retraining. Following the collector-diversity evidence~\cite{liber0}, collection should span many cooks and sites rather than many hours from a few people.

\paragraph{Acceptance.} Operators typically demand more than 95\% task success before committing~\cite{chefrobotics}. The following thresholds are proposed design targets, not a published standard or a validated deployment result. A staged rollout (test kitchen $\to$ one store off-peak $\to$ peak hours $\to$ multiple stores) with gates on success ($\geq$95\% then $\geq$98\%), takeovers ($\leq$2 then $\leq$1 per hour), availability ($\geq$98\%), cycle time ($\leq$1.5$\times$ then $\leq$1.2$\times$ a skilled worker), dosing error and zero food-safety incidents makes ``deployment-ready'' measurable. Subscription pricing, as used for kitchen packing robots~\cite{atombite}, lowers the upfront cost for restaurants. The exercise shows how deployment pushes the stack towards hybrid hand configurations, explicit contact layers, sensor-closed dosing and evaluation by takeovers and availability rather than success rate alone.

\section{Open Problems and Future Directions}
\label{sec:challenges}

We close the review with eleven open problems, each tied to the layers it spans.

\paragraph{P1. Durability under industrial duty cycles (L1).} Dividing a reported 300k-cycle rating by an illustrative 8k--10k grasps/day gives about 30--38 operating days, but this is a duty-cycle calculation, not a measured replacement interval; test loads and failure definitions may differ~\cite{yicaireliability}. The community needs load-specified, standardized endurance tests, field failure statistics, and designs for quick replacement of fingers, tendons and tactile skins~\cite{orcahand,dhrobotics}. Research directions include accelerated-life testing protocols with realistic grasp loads, wear-aware control that redistributes load across fingers, prognostics building on motor-current and tendon-tension sensing of contact and stiffness~\cite{hw_tendonprop,hw_dexhand021}, and designs that are robust to collisions by construction, as pursued long ago with antagonistic elastic drives~\cite{hw_awiwi}.

\paragraph{P2. Specification and reporting standards (L1, L5).} Total and active DoF, fingertip force, and cycle life are reported inconsistently. A minimal hardware datasheet (active DoF, coupling map, continuous and peak fingertip force, backdrivability, tactile resolution, endurance test conditions) would make results comparable across hands. Grasp-taxonomy coverage has saturated as a metric~\cite{hw_bidexhand,hw_dexlink}, so the community should agree on discriminative capability tests, for example combinations of Kapandji opposition, fingertip-force profiles, in-hand reorientation benchmarks and tool-use tasks.

\paragraph{P3. Co-designing hands with data (L1--L2).} Evidence that robots resembling humans benefit more from human data~\cite{liber0}, and that deployed-hand collection removes the gap~\cite{realdexumi}, suggests designing the hand, its cameras and its collection device together. How much anthropomorphism is needed, and for which tasks, is an open quantitative question~\cite{houseofdextra}, and the literature disagrees on whether anthropomorphism helps in-hand manipulation at all~\cite{hw_needanthro}. Controlled experiments that vary one design factor at a time (thumb range, palm articulation~\cite{hw_isyhand}, wrist DoF, finger length, camera placement) while holding data and models fixed would resolve much of this debate.

\paragraph{P4. Laws for data mixtures (L2).} Log-linear scaling has been observed for egocentric data up to about 20k hours~\cite{egoscale}, and collector diversity matters~\cite{liber0}, but there is no predictive theory of how to allocate a budget between egocentric, wearable, teleoperated and simulated data, or of when reconstruction quality rather than volume becomes the bottleneck. Controlled, open studies across multiple hands are needed, including the marginal value of an hour of each data type, the effect of reconstruction noise, and the interaction between data type and model size. The emergence of human-to-robot transfer only with sufficiently diverse robot pretraining~\cite{pih2r,hum_pihuman} suggests that these factors interact strongly.

\paragraph{P5. Tactile standardization (L1--L3).} Tactile sensors differ in principle, resolution and layout far more than joints differ in kinematics. Candidate canonical spaces include the MANO UV surface~\cite{unitachand} and unpaired latent alignment~\cite{tactalign,anytouch}, but cross-sensor transfer remains weak~\cite{touchscale}. Shared tactile data formats and paired multi-sensor datasets would accelerate progress, as would sensor-instance generalization of the kind demonstrated for replaceable magnetic skins~\cite{tac_anyskin} and cross-sensor pretraining at policy scale~\cite{tac_ftp1,tac_t3}. Tactile simulation needs validation protocols comparable to SIMPLER for vision~\cite{dep_simpler}.

\paragraph{P6. Zero-shot transfer between multi-fingered hands (L3).} Real zero-shot results above 80\% exist for grasping~\cite{dexgraspzero}, while transfer for in-hand manipulation and tool use is far less established, and much of the cross-embodiment evidence for generalist models comes from grippers~\cite{pi07,geminirobotics15,gen15}. Benchmarks that report zero-shot and $N$-shot adaptation curves across hands of very different morphology (e.g., a 6-DoF linkage hand, a 16-DoF four-finger hand and a 22-DoF anthropomorphic hand) would clarify where we stand.

\paragraph{P7. Action representations (L3--L4).} Relative EEF versus joint-space actions~\cite{gr00tn17,pi07}, explicit slots versus latent actions~\cite{unidex,opfa}, and pixel-space versus latent world-action models~\cite{dreamzero,beingh07} are all unresolved. Controlled ablations on dexterous (not only gripper) tasks are scarce. So are studies of action tokenization and inference speed at 20+ action dimensions and high control rates~\cite{pol_fast,pol_consistencypolicy}, where the cost of each design choice grows with the hand's DoF.

\paragraph{P8. Where contact control lives (L4).} Hierarchical systems put high-frequency contact control in a dedicated layer~\cite{figurehelix02,craftnet,agilinkduet}, while end-to-end models aim to learn it implicitly. Which split generalizes better, and how to train the interface between layers, are open questions. Academic dual systems without touch~\cite{pol_hirt,pol_robodual} and tactile slow--fast policies~\cite{tac_rdp,trex} provide components, but no public system yet couples a pretrained VLA with a learned tactile contact layer on a 20+-DoF hand and reports ablations of the interface. A related question is whether generated training data introduce ``looks right but cannot be done'' biases~\cite{robotryon,egoengine}.

\paragraph{P9. Evaluation of physical correctness (L5).} Completion rates hide failures of precision, force and damage, as Labimus shows for chemistry~\cite{labimus}. Benchmarks should report precision, applied force, object damage and human takeovers alongside success, use real-world counterparts with reported trial counts and confidence intervals, and validate that simulation scores predict real performance~\cite{teleopbench}.

\paragraph{P10. Safety, autonomy of evaluation and economics (L5).} Safety standards lack hand-specific pinch and crush limits~\cite{dep_biomechlimits,dep_painthreshold}; automatic resets and success detection exist for tabletop grippers~\cite{dep_autoeval} but not for in-hand tasks; and there is no peer-reviewed evidence on the economics of dexterous deployment. All three hinder the transition from demonstrations to products.

\paragraph{P11. Governance of human data (L2).} Large-scale egocentric and glove capture in homes and workplaces raises privacy, consent and labor questions~\cite{techreviewgig,humanarchive}. Emerging dataset standards~\cite{miitstandard,isostandard} provide quality criteria, but provenance tracking, consent management and de-identification for manipulation data are still immature.

\subsection{Cross-layer conclusions and decisive tests}
\label{sec:synthesis}

A survey should distinguish repeated observations from assumptions that still need a controlled test. \Cref{tab:synthesis} links the principal conclusions to their evidence and limits. The common theme is conditionality: alignment can make human data useful, but its value depends on embodiment; contact feedback can improve difficult tasks, but the best architectural split is unresolved; and a high completion score can coexist with unacceptable force, damage or intervention rates.

\begin{table*}[t]
\centering\small
\setlength{\tabcolsep}{4pt}
\caption{Cross-layer synthesis: what the evidence supports, what remains unresolved, and which comparison would be decisive. Tests in the final column are proposed experiments.}
\label{tab:synthesis}
\begin{tabularx}{\textwidth}{L{2.5cm}YYY}
\toprule
Claim & Supporting evidence & Boundary or counterpoint & Decisive next comparison \\
\midrule
Aligned human data can improve learning & EgoScale and DexWild report benefits of human pretraining/co-training~\cite{egoscale,dexwild}. & More human data is not always better; gains depend on the bridge and mixture. & Vary source hours and alignment quality at fixed compute, hand and robot data. \\
\addlinespace
Contact should constrain transfer & Physics and contact-aware methods improve transferred trajectories~\cite{spider,contactanchored}. & Simulated contact feasibility does not guarantee real friction or force accuracy. & Match human clips and target hands; compare kinematic, contact and real-task errors. \\
\addlinespace
No action space is universally best & Relative EEF and joint-space choices receive different support~\cite{gr00tgithub,pi07}. & Embodiment sets, tasks and training pipelines differ across reports. & Compare action spaces on the same data, backbone and morphologically distant hands. \\
\addlinespace
Touch and fast feedback are promising & Tactile mid-training and hierarchies show task-specific gains~\cite{touchscale,trex}. & Sensor heterogeneity persists; a separate contact layer is not always tested against end-to-end control. & Factor sensing, controller split and latency independently; evaluate unseen sensors. \\
\addlinespace
Zero-shot dexterity remains narrow & Real cross-hand grasp transfer has been reported~\cite{dexgraspzero}. & Grasp transfer does not establish long-horizon tool use or in-hand skill transfer. & Report zero-shot and fixed-budget adaptation on grasping, reorientation and tool use. \\
\addlinespace
Completion is not deployment readiness & Laboratory tasks expose divergence between completion and validity~\cite{labimus}. & Uptime, repair, damage and cost evidence is sparse or company-reported. & Use fixed-duration trials with intervention, failure and maintenance logs. \\
\bottomrule
\end{tabularx}
\end{table*}

\subsection*{Recommendations by stakeholder}

\begin{table*}[t]
\centering\tabsize
\caption{Recommendations by stakeholder, derived from the evidence reviewed in this survey.}
\label{tab:stakeholders}
\begin{tabularx}{\textwidth}{L{2.4cm}YY}
\toprule
Stakeholder & What the evidence supports & What to measure and report \\
\midrule
Hand designers & Backdrivable joints and low backlash for learning; human-like proportions, wrist and camera placement where human data are used; hybrid actuation for force plus precision; modular, quick-swap fingers, tendons and skins & Active DoF and coupling map; continuous and peak fingertip force; endurance under specified load; tactile resolution and coverage; repair time \\
Data teams & Store raw human and robot signals with calibration; mirror camera placement; isomorphic devices with shared touch; diverse collectors and sites; online quality checks; consent and de-identification from day one & Valid-data rate; synchronization error; collector and site diversity; cost per accepted hour; provenance \\
Cross-embodiment researchers & Contact-preserving retargeting; slot action spaces with masks; paired human--robot data; tests on morphologically distant hands & Zero-shot and $N$-shot curves across at least two hand types; contact error, not only joint error \\
Model developers & Large-scale pretraining plus paired mid-training; touch as an input with gating or a separate fast layer; RL and interventions after imitation; fast inference for high-rate control & Success with confidence intervals; latency and control rate; behavior under perturbation; data scale per stage \\
Benchmark designers & Functional, tactile and long-horizon tasks; embodiment-decoupled task definitions; real-world counterparts & Precision, force, damage and takeovers alongside success; sim--real correlation; trial counts \\
Deployers & Use dexterous hands only where needed; deterministic safety layers; sensor-closed dosing; staged rollout with gates & Task success, takeovers per hour, availability, cycle time vs.\ skilled worker, MTBF, incident counts \\
\bottomrule
\end{tabularx}
\end{table*}

\Cref{tab:stakeholders} summarizes recommendations by stakeholder.

\paragraph{A minimal recipe.} For practitioners building a dexterous stack today, the evidence suggests: (1)~store raw sources (MANO keypoints or glove angles, 6D wrist poses, raw video and depth, touch, camera calibration and object meshes and poses) and treat retargeted actions as recomputable derivatives; (2)~adopt a function-aligned slot action space with masks, and run an explicit ablation of relative-EEF versus joint-space wrist actions on your own embodiments; (3)~make retargeting contact-preserving and filter converted data with physics replay; (4)~plan for tens of hours of paired human--robot data and a few hours of target-hand data, and start adaptation experiments at the ``30 minutes or ten demonstrations'' scale suggested by recent results~\cite{dreamzero,opfa}; (5)~include touch in the data specification from day one; and (6)~evaluate across at least two hands of different morphology.

\FloatBarrier
\section{Conclusion}
\label{sec:conclusion}

Progress in dexterous robotics is best understood through the interfaces between hardware, data, representations, policies and evaluation. Lower-cost hands and larger human-data corpora expand the design space, but neither price nor recording hours alone predicts usable dexterity. The reviewed within-study comparisons support the value of alignment, contact information and targeted robot adaptation under their respective protocols. They do not establish a universal data mixture, action space or control hierarchy.

Three priorities follow. First, retain raw observations, calibration, action semantics and embodiment metadata so that transfer errors can be diagnosed and corrected. Second, compare shared foundation policies and embodiment-specific contact controllers under matched sensing, compute and latency, including hand types that differ substantially in morphology. Third, evaluate physical correctness and long-duration reliability alongside completion, with trial counts, intervention budgets and explicit failure definitions. The strongest unresolved claims concern zero-shot contact-rich transfer across hands, cross-sensor tactile generalization, and sustained deployment economics. Addressing these gaps with controlled experiments and transparent reporting would turn a rapidly expanding catalog of systems into a cumulative science of dexterity.

\bibliographystyle{unsrtnat}
\bibliography{bib/refs}

\end{document}